# AIPatient Arena: EHR-grounded evaluation of large language models in end-to-end clinical consultation workflows

Jiahui Niu[1,2§], Huizi Yu[3§], Wenkong Wang [1,2§], Guangxin Dai [1,2], Jingxian He[4], Xiang Li[5], Zhiying Liang[5], Xinxin Lin[5], Kent CY So[3], Bryan YP Yan[3], Yun Kwok Wing[5,6,7,8], Yanqiu Xing[4*], Xin Ma[1,2*], Lizhou Fan[5,7*]

1. School of Control Science and Engineering, Shandong University, Ji'nan, Shandong, China
2. Key Laboratory of Machine Intelligence and System Control, Shandong University, Ji'nan, Shandong, China
3. Department of Medicine and Therapeutics, The Chinese University of Hong Kong, Shatin, Hong Kong SAR, China
4. Department of Geriatric Medicine, Qilu Hospital of Shandong University, Ji'nan, Shandong, China
5. Department of Psychiatry, The Chinese University of Hong Kong, Shatin, Hong Kong SAR, China
6. Li Chiu Kong Family Sleep Assessment Unit, Department of Psychiatry, Faculty of Medicine, The Chinese University of Hong Kong, Shatin, Hong Kong SAR, China
7. Li Ka Shing Institute of Health Sciences, Faculty of Medicine, The Chinese University of Hong Kong, Shatin, N.T., Hong Kong SAR, China
8. Gerald Choa Neuroscience Institute, Department of Medicine and Therapeutics, The Chinese University of Hong Kong, Hong Kong SAR, China

[§] Jiahui Niu, Huizi Yu, and Wenkong Wang contributed equally.

* Correspondence to Lizhou Fan (leofan@cuhk.edu.hk), Xin Ma (maxin@sdu.edu.cn), and Yanqiu Xing (xingyanqiu@qiluhospital.cn).

## Abstract

Large language models (LLMs) are increasingly considered for use in clinical consultation tasks, yet most medical evaluations remain static, single-turn, or narrowly outcome-based, limiting their ability to reflect the sequential, uncertain, and interactive nature of real-world care, such as those documented in electronic health records (EHR). Here, we propose AIPatient Arena, an EHRs-grounded evaluation framework for assessing the clinical utility of LLMs in consultation across eight dimensions of clinical competence. The framework integrates EHR data into patient-specific knowledge graphs, enabling multi-turn physician–patient interactions. We applied the AIPatient Arena evaluation framework on a primary cohort of 437 patients and two out-of-distribution external validation cohorts of 119 and 67 patients respectively. We observe that contemporary LLMs performed relatively well in medical interview questioning skills (QS; mean scores, 4.43–4.99/5), ethical and professional conduct (ET; 4.38–4.93/5), and clarity and transparency of clinical explanations (EX; 3.80–4.72/5). Performance was moderate in information integration (II; 3.19–4.21/5) and medication safety and justification (MS; 3.13–3.78/5), but persistent weaknesses were observed in clinically consequential dimensions, particularly handling of ambiguous patient responses (HR; 2.57–3.32/5), information coverage (IC; 2.08–3.02/5), and diagnostic accuracy and reasoning (Dx; 2.63–3.55/5). Process-based evaluation revealed recurrent interaction failures, including repetitive questioning, omission of past medical history, inadequate handling of uncertainty, and limited empathy. Richer conversational context improved diagnostic reasoning but yielded limited gains in treatment planning. These findings indicate that final-answer accuracy alone is insufficient for evaluating clinical readiness and highlight the importance of assessing how models gather, interpret, and communicate information throughout a consultation. AIPatient Arena provides an EHR-grounded framework for workflow-oriented pre-deployment evaluation of medical LLMs.

## Introduction

Large language models (LLMs) are increasingly being explored for clinical applications, including documentation assistance, decision support, risk assessment, patient communication, and medical information synthesis[1–4]. Yet the gap between the seemingly strong performance in current benchmarks and practical clinical use remains substantial. While sometimes LLM outputs appear to be fluent and clinically plausible, there are remaining concerns about their readiness for real-world use in patient-facing settings[5], due to the often observed omissions, unsupported inferences, and unsafe recommendations. Real-world clinical consultations are particularly demanding because they require not only medical knowledge, but also structured questioning, clarification of ambiguous responses, integration of incomplete information, transparent explanation, and treatment reasoning under uncertainty[6,7].

Most existing benchmarks of medical LLMs do not fully capture real-world clinical consultation complexity. Benchmarks for medical and healthcare models are often developed with static examination-style, especially multiple-choice medical question answering, which could have overestimated clinical readiness by rewarding recognition of pre-specified facts rather than interactive reasoning in realistic workflows[8–10]. Recent work has begun to address this limitation. CRAFT-MD[11] is a benchmark using simulated doctor–patient dialogue for evaluating conversational diagnostic reasoning and showed that model performance deteriorates when moving from static vignettes to conversational settings. MedHELM[6] substantially broadened evaluation by organizing medical AI applications into a clinician-validated taxonomy spanning clinical decision support, note generation, patient communication, research, and administration. However, important gaps remain. Existing frameworks still provide limited evaluation of whether an LLM can participate safely and effectively in an end-to-end clinical consultation, at which the model must gather information over time, respond appropriately to uncertainty, and justify diagnostic and treatment decisions in a clinically coherent and patient-understandable manner. At the same time, there is relative scarcity of evaluation frameworks grounded in real-world electronic health record (EHR) data. EHRs contain the ambiguity, incompleteness, shorthand, and contextual heterogeneity that characterize real-world clinical practice. These real-world challenges call for model robustness and adaptability.

In this study, we present AIPatient Arena, an EHR-grounded evaluation framework for assessing LLM performance in simulated end-to-end clinical consultations. Built on our previous work in AIPatient simulation framework[12], which offered potential in training medical students, and grounded in Medical Information Mart for Intensive Care (MIMIC)-III[13] critical care EHRs, AIPatient Arena constructs patient-specific knowledge graphs to support realistic multi-turn physician–patient interactions. Within these interactions, the evaluated LLM assumes the role of the physician and is expected to perform four core clinical functions: clinical communication, information summarization, diagnosis generation, and treatment planning. Essentially, our proposed framework evaluates LLMs in the same way of assessing medical student, aiming to prepare the models for real-world clinical consultation complexity.

AIPatient Arena evaluates model performance across eight clinically grounded dimensions co-designed by attending physicians and medical AI experts: medical interview questioning skills (QS), information coverage (IC), handling of ambiguous patient responses (HR), ethical and professional conduct (ET), clarity and transparency of clinical explanations (EX), information integration (II), diagnostic accuracy and reasoning (Dx), and medication safety and justification (MS). Each dimension includes a set of predefined failure patterns that capture clinically meaningful weaknesses in model behavior, identified by experienced physicians, such as repetitive questioning, omission of past medical history, failure to address uncertain answers, and lack of empathy. During evaluation, occurrences of the failure patterns are identified and counted, and dimension-specific scores are deducted according to a rubric-based penalty scheme.

This failure-pattern-based scoring design enables process-level assessment beyond final diagnostic or treatment accuracy and allows clinically interpretable localization of LLM weaknesses across the consultation pathway. Using this framework, we benchmark contemporary general-purpose LLMs, including commercial and open source ones, and post-trained LLMs for the medicine domain, including GPT-5.5[14], GPT-5[15], GPT-4o[16], Claude 4.6 Sonnet[17], Claude 4.0 Sonnet[18], DeepSeek-V4-Pro[19], DeepSeek-V3[20], Qwen3.5-397B[21], Qwen3-235B[22], MedGemma-27B-Text-IT[23], HuatuoGPT-o1-72B[24], and Baichuan-M2-32B[25]. We also evaluate the robustness and generalizability of the AIPatient Arena framework across models with two out-of-distribution clinical cohorts.

Overall, the AIPatient Arena provides an EHR-grounded, multidimensional, and human-centered benchmark framework for evaluating LLMs before they can be potentially deployed in real-world clinical consultation settings, providing a complementary layer of evaluation beyond of the current static and knowledge-based evaluation of LLMs in medicine.

## Results

### AIPatient Arena enables EHR-grounded evaluation of end-to-end clinical consultations

The AIPatient Arena framework was designed to evaluate large language models under clinically realistic consultation workflows by grounding simulated patient interactions in heterogeneous EHR data sources. Using a combination of structured EHR processing and language model-based clinical information extraction, we constructed patient-specific knowledge representations that integrated symptoms, medical history, medications, laboratory findings, and other clinically relevant information across longitudinal patient records (**Fig. 1a**). Analyses were conducted on the Clinical Consultation Question-Answering (CCQA) cohort from MIMIC-III[13], comprising 442 admissions from 437 adult patients. The resulting patient graph contained 9,682 nodes and 19,752 relationships, with more than half of nodes originating from narrative clinical documentation. By grounding the simulated consultation in both structured EHR records and narrative clinical text, AIPatient Arena is designed to better approximate the informational complexity of real-world clinical encounters than vignette-based testing environments.

These patient-specific knowledge representations were further used to construct AIPatient for end-to-end consultation simulation. During each consultation, LLMs iteratively gathered clinically relevant information from the AIPatient under a constrained consultation horizon before generating outputs for information summarization, diagnosis generation, and treatment planning (**Fig. 1b**). This design enabled evaluation of whether the model could effectively prioritize questioning, identify clinically relevant information, and maintain coherent clinical reasoning throughout sequential patient interactions.

Consultation quality was assessed across eight clinically grounded dimensions spanning medical interview questioning skills, information coverage, handling of ambiguous patient responses, ethical and professional conduct, clarity and transparency of clinical explanations, information integration, diagnostic accuracy and reasoning, and medication safety and justification. Each dimension was associated with predefined failure patterns that were used by an AI evaluator to identify consultation deficiencies and apply score deductions to LLM outputs, with the evaluation process verified by clinicians (**Fig. 1c**). Rather than reducing consultation quality to a single endpoint, this multidimensional structure allows the framework to capture distinct components of clinical performance that are directly relevant to safe and effective patient-facing use.

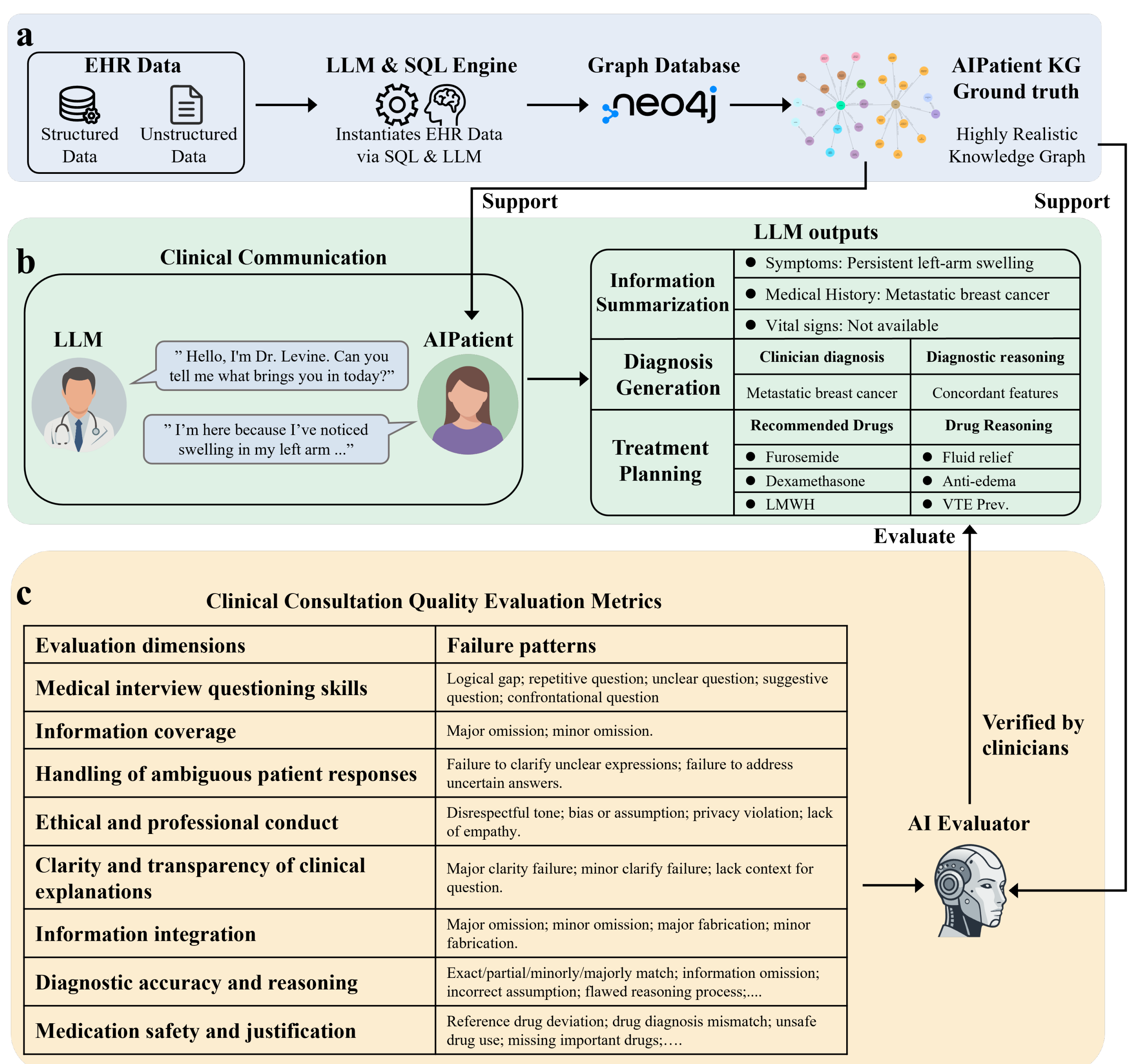

| Evaluation dimensions | Failure patterns |
|---|---|
| Medical interview questioning skills | Logical gap; repetitive question; unclear question; suggestive question; confrontational question |
| Information coverage | Major omission; minor omission. |
| Handling of ambiguous patient responses | Failure to clarify unclear expressions; failure to address uncertain answers. |
| Ethical and professional conduct | Disrespectful tone; bias or assumption; privacy violation; lack of empathy. |
| Clarity and transparency of clinical explanations | Major clarity failure; minor clarify failure; lack context for question. |
| Information integration | Major omission; minor omission; major fabrication; minor fabrication. |
| Diagnostic accuracy and reasoning | Exact/partial/minorly/majorly match; information omission; incorrect assumption; flawed reasoning process;.... |
| Medication safety and justification | Reference drug deviation; drug diagnosis mismatch; unsafe drug use; missing important drugs;.... |

**Fig. 1 | AIPatient Arena: an EHR-grounded framework for evaluating large language models in end-to-end clinical consultations. a**, EHR-grounded patient construction. Structured EHR data are processed using SQL pipelines, while entities are extracted from unstructured clinical narratives using large language models. These heterogeneous clinical data are then integrated into a patient-specific Neo4j knowledge graph, which serves as the ground truth for consultation simulation and evaluation. **b**, Simulation of end-to-end consultations. An LLM interacts with the AIPatient through a multi-turn consultation, asking one question at a time to elicit symptoms, prior history, and other clinically relevant information. The interaction is constrained by a finite consultation horizon to assess information prioritization and consultation discipline under realistic workflow limits. At the end of the consultation, LLM outputs for information summarization, diagnosis generation, and treatment planning. **c**, Multidimensional evaluation of consultation quality. Following each consultation, LLM outputs are evaluated across eight clinically grounded dimensions, including medical interview questioning skills, information coverage, handling of ambiguous patient responses, ethical and professional conduct, clarity and transparency of clinical explanations, information integration, diagnostic accuracy and reasoning, and medication safety and justification. Each dimension contains multiple predefined failure patterns, and failures identified in the LLM outputs lead to corresponding score deductions according to a rubric-based scoring scheme, with the evaluation process verified by clinicians.

## Human review supports the validity of multidimensional consultation assessment

To examine the alignment between automated evaluation scores and human clinical judgment, four large language models (GPT-4o[16], Claude 4.0 Sonnet[18], Qwen3[22], and MedGemma[23]) generated responses across four dialogue datasets, resulting in 120 simulated clinical consultations. Each consultation was independently evaluated by five medical experts. Human review supported the overall validity of this multidimensional evaluation design. Across the four datasets, expert evaluations showed broad consistency with the automated clinical assessment framework. Median satisfaction scores exceeded 4.20 on a 5-point Likert scale in all 32 evaluated scenarios (**Supplementary Table 1**).

Among the four LLMs, with MedGemma-generated consultations (**Supplementary Fig. 1d**) received the highest expert ratings, with median scores reaching 4.95. Across interaction-centric evaluation dimensions, including QS, EX and ET, the automated evaluator produced highly concentrated score distributions. For MedGemma and Claude 4.0 Sonnet (**Supplementary Fig. 1b**), these dimensions exhibited particularly narrow interquartile ranges (IQRs), with values as low as 0.026 and 0.082, indicating minimal variability across consultations. While Qwen3-generated consultations (**Supplementary Fig. 1c**) showed greater dispersion in QS, with an IQR of 0.272, while the overall score distributions remained centered at high median values (median QS = 4.48).

A characteristic expansion in score variance was consistently observed in core clinical reasoning dimensions, specifically Dx and MS. Across all four model-generated datasets, scores in these dimensions exhibited the broadest distributions; for instance, medication safety scores in the GPT-4o cohort (**Supplementary Fig. 1a**) spanned from 3.56 to 4.60 (IQR = 0.29). This downward shift and elongation of the boxplots capture the inherent rigor and professional subjectivity associated with high-stakes medical decision-making. Despite the increased dispersion, which reflects the stringent criteria applied by experts to pharmacological and diagnostic interventions, median scores remained above 4.20 for GPT-4o and reached as high as 4.64 for MedGemma, underscoring the framework's capacity to provide nuanced assessments of AI-mediated clinical reasoning.

## LLMs perform better on interactional behaviors than on reasoning and medication safety

AIPatient Arena quantifies clinical consultation quality across eight predefined dimensions encompassing communication, reasoning, and safety-related competencies. We reported the dimension-level scores for each model, together with an overall weighted average that aggregates performance according to predefined clinical importance weights (**Fig. 2** and **Table 1**).

Across LLMs, performance was generally stronger on interactional and communication-related aspects of the consultation than on clinically consequential reasoning and treatment-related dimensions. Models frequently received relatively high scores for QS, ET, and EX, all of which approached near-ceiling performance, indicating that they were often able to sustain plausible multi-turn conversations and demonstrate appropriate professional behavior. Performance in II and MS was comparatively more moderate, with scores commonly ranging between 3 and 4. By contrast, broader variability and lower performance were observed in IC, HR, and Dx. DeepSeek-V3 scored as low as 2.08 in IC, while the highest HR score was only 3.32, achieved by HuatuoGPT-o1. For Dx, only the GPT-5 series achieved relatively higher scores (GPT-5.5[14]: 3.55; GPT-5[15]: 3.37). Collectively, these findings suggest a dissociation between conversational fluency and clinically reliable downstream decision-making.

Regarding the comparative performance of the evaluated LLMs across the eight clinically relevant dimensions, substantial differences were observed in both overall capability and dimensional performance profiles. GPT-5.5 achieved the highest weighted average score of 3.76 among all LLMs, with its advantage reflecting more balanced performance across clinically critical dimensions rather than exceptional performance in a single category. In particular, GPT-5.5 attained the highest scores in QS (4.99), II (4.21), and Dx (3.55), although performance in IC remained limited (2.48). GPT-5 ranked second overall with a weighted average score of 3.69, while Qwen3.5[21] and DeepSeek-V4-Pro[19] followed closely (3.64 and 3.63, respectively), both demonstrating strong performance in MS, including the highest MS score achieved by Qwen3.5 (3.78). HuatuoGPT-o1[24] and Baichuan M2[25] demonstrated intermediate overall performance (3.57 and 3.53), with HuatuoGPT-o1 achieving the highest score in HR (3.32). Claude 4.6 Sonnet[17] and Claude 4.0 Sonnet showed similar overall performance (3.51 each); notably, Claude 4.0 Sonnet was the only model to surpass a score of 3 in IC, while also maintaining near-ceiling performance in ET (4.93). GPT-4o, DeepSeek-V3, and MedGemma exhibited closely aligned performance profiles, with weighted average scores clustered between 3.41 and 3.47, although DeepSeek-V3 achieved the highest score in EX. In contrast, Qwen3 demonstrated greater variability across evaluation dimensions and lower overall performance (3.36). Overall, none of the LLMs demonstrated consistently strong performance across all eight dimensions, highlighting persistent trade-offs between communication-related abilities and clinically consequential reasoning tasks. Together, these findings demonstrate the value of multidimensional evaluation for characterizing clinically relevant strengths and limitations across different LLMs.

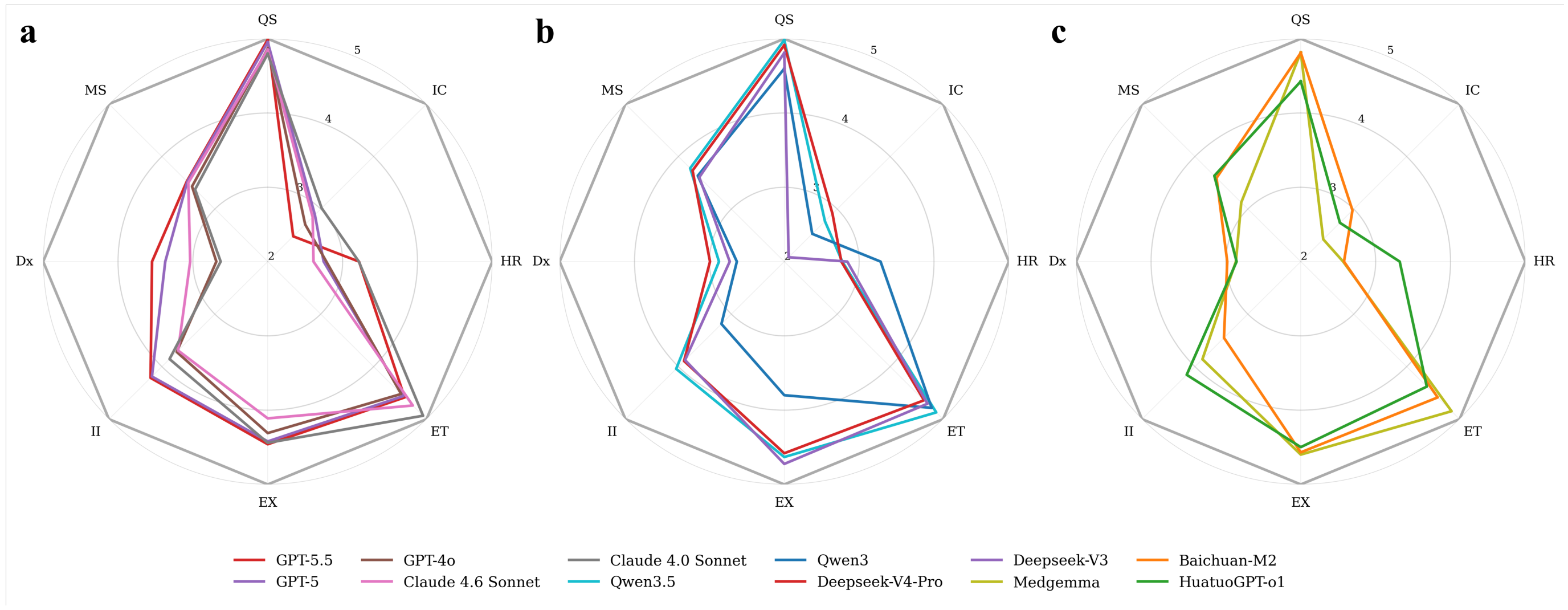

**Fig. 2 | Comparative performance of LLMs across clinical evaluation dimensions.** We compared models in different categories, including **a**, Closed-source commercial LLMs, **b**, Open-source LLMs, and **c**, Medical-specialized LLMs. Models demonstrate stronger performance in communication-oriented skills, particularly QS, ET and EX, while exhibiting relatively weaker performance in IC, HR and Dx.

**Table 1 | Performance of LLMs across the eight evaluation dimensions.** GPT-5.5 achieved the highest overall score (3.76), with different models attaining the top score in individual dimensions**.**

| Model | Size | QS | IC | HR | ET | EX | II | Dx | MS | Wtd. Avg. |
|---|---|---|---|---|---|---|---|---|---|---|
| Claude 4.0 Sonnet | - | 4.80 | **3.02** | 3.22 | **4.93** | 4.44 | 3.86 | 2.63 | 3.37 | 3.51 |
| Claude 4.6 Sonnet | - | 4.85 | 2.85 | 2.61 | 4.74 | 4.11 | 3.69 | 3.04 | 3.50 | 3.51 |
| GPT-4o | - | 4.87 | 2.71 | 2.78 | 4.52 | 4.31 | 3.72 | 2.69 | 3.43 | 3.42 |
| GPT-5 | - | 4.95 | 2.89 | 2.75 | 4.56 | 4.42 | 4.19 | 3.37 | 3.52 | 3.69 |
| GPT-5.5 | - | **4.99** | 2.48 | 3.22 | 4.59 | 4.46 | **4.21** | **3.55** | 3.53 | **3.76** |
| Deepseek-V3 | 671B | 4.81 | 2.08 | 2.84 | 4.71 | **4.72** | 3.87 | 2.73 | 3.61 | 3.47 |
| Deepseek-V4-Pro | 1.6T | 4.91 | 2.98 | 2.76 | 4.64 | 4.58 | 3.89 | 2.99 | 3.73 | 3.63 |
| Qwen3-235B | 235B | 4.60 | 2.53 | 3.29 | 4.79 | 3.80 | 3.19 | 2.63 | 3.63 | 3.36 |
| Qwen3.5-397B | 397B | 4.98 | 2.77 | 2.77 | 4.87 | 4.63 | 4.04 | 2.87 | **3.78** | 3.64 |
| MedGemma-27B-Text | 27B | 4.82 | 2.42 | 2.57 | 4.85 | 4.60 | 3.86 | 2.87 | 3.13 | 3.41 |
| Baichuan-M2-32B | 32B | 4.81 | 2.98 | 2.58 | 4.59 | 4.57 | 3.45 | 2.98 | 3.59 | 3.53 |
| HuatuoGPT-o1 | 72B | 4.43 | 2.74 | **3.32** | 4.38 | 4.50 | 4.16 | 2.86 | 3.63 | 3.57 |

To further examine relationships among the evaluation dimensions, we analyzed pairwise correlations using pooled case-level scores across all models. As shown in **Supplementary Fig. 2**, correlations between dimensions were uniformly weak, with all absolute values below 0.13 and no moderate or strong associations observed, indicating that each dimension captures a distinct aspect of model behavior and supports their use as complementary components of the evaluation framework.

### Failure patterns in dialogue-level evaluation dimensions

We analyzed four dialogue-level evaluation dimensions, each of which encompasses multiple fine-grained failure patterns (**Fig. 3** and **Table 2**). These dimensions captured key components of clinical interaction, including contextual continuity, information coverage, responses to ambiguous patient statements and ethical and professional conduct.

For information coverage, omission of past medical history represented the predominant failure pattern across most models. DeepSeek-V3 showed the highest past-medical-history omission rate for this category (0.952), followed by GPT-5.5 (0.928), GPT-5 (0.871), Qwen3 (0.835) and Qwen3.5 (0.826). The remaining models had lower but still substantial omission rates, ranging from 0.618 for HuatuoGPT-o1 to 0.692 for MedGemma. These results suggest that incomplete history taking was a common limitation across model families rather than a failure restricted to a small subset of models.

The handling of ambiguous patient responses was dominated by failure to address uncertain answers, highlighting a systematic inability of LLMs to address unclear patient replies. Failure rates exceeded 0.9 for nearly all models, with Deepseek-V4-Pro exhibiting the highest rate (0.991), suggesting that this failure pattern occurred in almost every case.

A uniform pattern was observed in ethical and professional conduct dimension, where lack of empathy consistently was the most prevalent failure pattern across all LLMs. HuatuoGPT-o1 showed a relatively high lack-of-empathy rate (0.455), while GPT-4o and GPT-5 also exhibited elevated rates (0.387 and 0.348, respectively). In contrast, the comparatively smaller MedGemma achieved a lower failure rate of 0.118, and Claude 4.0 Sonnet and Claude 4.6 Sonnet showed the lowest rates, both at 0.048.

In medical interview questioning skill, repetitive questioning was the most common pattern, reflecting a potential model failure to effectively incorporate contextual information and a tendency to revisit previously elicited content. While reframing the questions in real-world clinical setting might add value to patient-facing communication and clinical information exploration, this repetitiveness for facing AIPatient-based simulated responses is rather a system redundancy issue. Here, HuatuoGPT-o1 had the highest repetitive-questioning rate (0.520), followed by Qwen3 (0.382), DeepSeek-V3 (0.262), Baichuan-M2 (0.249) and MedGemma (0.242). In comparison, several closed-source models showed lower rates, with Claude 4.0 Sonnet and GPT-4o performing relatively well and GPT-5.5 achieving the lowest rate (0.011). However, low repetitive-questioning rates were not limited to closed-source models, as DeepSeek-V4-Pro (0.063) and Qwen3.5 (0.018) also performed well on this pattern.

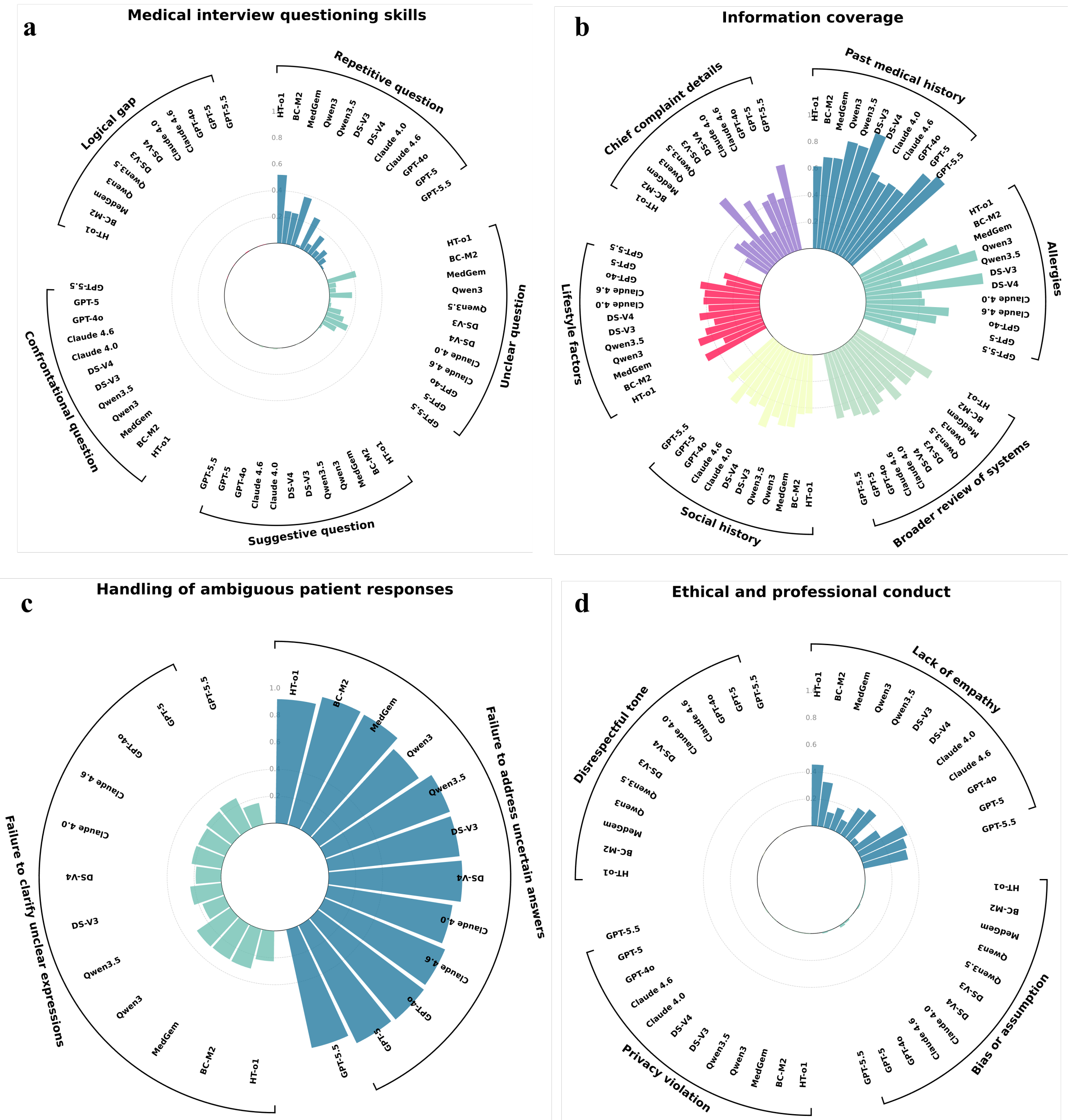

**Fig. 3 | Failure patterns and performance landscapes across dialogue-level evaluation dimensions. a,** Medical interview questioning skills: LLMs occasionally produce repetitive or unclear questions during clinical communication, with HuatuoGPT-o1 showing the highest failure probability. **b,** Information coverage: LLMs frequently demonstrate inadequate coverage in clinical interviews, especially regarding past medical history and allergies. **c,** Handing of ambiguous patient responses: LLMs tend to respond superficially to unclear patient answers and often fail to follow up to obtain clarification. **d,** Ethical and professional conduct: LLMs demonstrate basic adherence to medical ethics, but lack sufficient empathy in patient interactions.

**Table 2 | LLM performance in predominant failure patterns across dialogue-level evaluation dimensions.** HuatuoGPT-o1 exhibited the highest probabilities in the failure patterns of repetitive questions and lack of empathy, with values of 0.520 and 0.455, respectively. DeepSeek-V3 showed a markedly high probability (0.952) of omitting inquiries about past medical history. In the uncertain answer handling category, all LLMs demonstrated consistently high failure rates, exceeding 90% in nearly all cases.

| Model | Repetitive questions | Past medical history | Uncertain answer handing | Lack of empathy |
|---|---|---|---|---|
| Claude 4.0 Sonnet | 0.163 | 0.631 | 0.934 | 0.048 |
| Claude 4.6 Sonnet | 0.057 | 0.658 | 0.973 | 0.048 |
| GPT-4o | 0.100 | 0.656 | 0.986 | 0.387 |
| GPT-5 | 0.057 | 0.871 | 0.975 | 0.348 |
| GPT-5.5 | 0.011 | 0.928 | 0.898 | 0.335 |
| Deepseek-V3 | 0.262 | **0.952** | 0.980 | 0.238 |
| Deepseek-V4-Pro | 0.063 | 0.672 | **0.991** | 0.278 |
| Qwen3-235B | 0.382 | 0.835 | 0.896 | 0.170 |
| Qwen3.5-397B | 0.018 | 0.826 | 0.986 | 0.102 |
| MedGemma-27B-Text | 0.242 | 0.692 | 0.975 | 0.118 |
| Baichuan-M2-32B | 0.249 | 0.690 | 0.982 | 0.333 |
| HuatuoGPT-o1 | **0.520** | 0.618 | 0.921 | **0.455** |

Overall, the dialogue-level evaluation identified multiple limitations in the consultation capabilities of LLMs. Repetitive questioning suggested insufficient contextual integration, whereas omission of past medical history reflected incomplete clinical information gathering. Failure to address uncertain answers indicated difficulties in handling ambiguity, and lack of empathy highlighted weaknesses in patient-centred communication. These observations emphasize that evaluation of clinical LLMs should extend beyond diagnostic accuracy to include the quality of the consultation process itself.

### Dialogue-sufficient consultations show higher diagnostic performance but limited changes in medication safety

In clinical consultations, failures in downstream diagnostic reasoning or treatment planning may arise not only from deficiencies in model reasoning, but also from insufficient information elicited during the dialogue. To disentangle these effects, the proposed evaluation framework explicitly assesses whether a model–patient interaction provides sufficient clinical information to support downstream decision-making.

For each model, encounters meeting this dialogue-sufficiency criterion were first identified. We include 190 dialogue-sufficient encounters where diagnosis may need further information beyond the initial round of clinical consultation. We report Dx and MS scores computed on this subset **Fig. 4** and **Table 3**).

All models achieved higher Dx scores in dialogue-sufficient cases than in all cases, with ΔDx ranging from 0.15 (Baichuan-M2) to 0.28 (Deepseek-V3). GPT-5.5 achieved the highest Dx score under dialogue-sufficient conditions (3.81), followed by GPT-5 (3.62), Claude 4.6 Sonnet (3.28), Deepseek-V4-Pro (3.16), Baichuan-M2 (3.13), Qwen3.5 (3.11), MedGemma (3.09), HuatuoGPT-o1 (3.07) and Deepseek-V3 (3.01). GPT-4o (2.90) and Claude 4.0 Sonnet (2.89) exhibited comparable Dx performance, while Qwen3 showed the lowest Dx score among LLMs (2.82).

In contrast, models showed minimal changes in MS scores, with ΔMS values close to zero and ranging from -0.02 (Qwen3) to 0.08 (Claude 4.0 Sonnet). Qwen3.5 achieved the highest MS score under dialogue-sufficient conditions (3.84). Deepseek-V4-Pro, HuatuoGPT-o1, Qwen3, DeepSeek-V3, Baichuan-M2, GPT-5, GPT-5.5, Claude 4.6 Sonnet, GPT-4o, and Claude 4.0 Sonnet achieved comparable scores (3.45-3.79), whereas MedGemma had the lowest score (3.15).

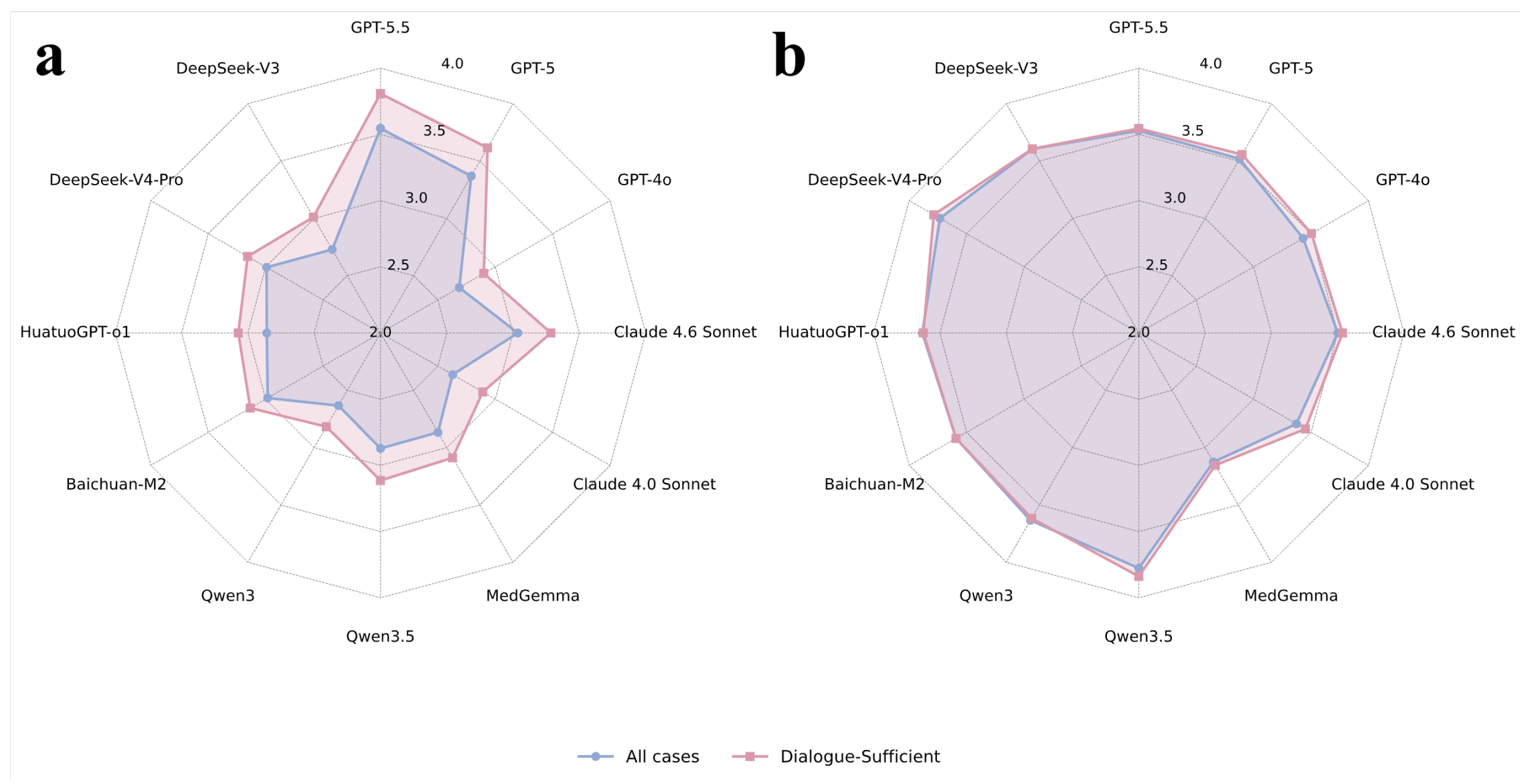

**Fig. 4 | All vs Dialogue-sufficient performance on Dx and MS metrics. a,** Model performance on Dx. Restricting the analysis to dialogue-sufficient cases consistently improved Dx scores across all models. **b,** Model performance on MS. Restricting the analysis to dialogue-sufficient cases did not produce noticeable changes in MS scores.

**Table 3 | LLM performance on Dx and MS in dialogue-sufficient encounters.** Dx scores ranged from 2.82 to 3.81, with score differences (ΔDx) ranging from 0.15 to 0.28. MS scores ranged from 3.15 to 3.84, with score differences (ΔMS) ranging from −0.02 to 0.08.

| Model | Count | Intersection N | Dx | MS | ΔDx | ΔMS |
|---|---|---|---|---|---|---|
| Claude 4.0 Sonnet | 314 | 190 | 2.89 | 3.45 | 0.26 | **0.08** |
| Claude 4.6 Sonnet | 319 | 190 | 3.28 | 3.54 | 0.25 | 0.04 |
| GPT-4o | 319 | 190 | 2.90 | 3.50 | 0.21 | 0.07 |
| GPT-5 | 320 | 190 | 3.62 | 3.56 | 0.25 | 0.04 |
| GPT-5.5 | 307 | 190 | **3.81** | 3.55 | 0.26 | 0.02 |
| Deepseek-V3 | 299 | 190 | 3.01 | 3.61 | **0.28** | 0.00 |
| Deepseek-V4-Pro | 314 | 190 | 3.16 | 3.79 | 0.16 | 0.05 |
| Qwen3-235B | 298 | 190 | 2.82 | 3.62 | 0.18 | -0.02 |
| Qwen3.5-397B | 314 | 190 | 3.11 | **3.84** | 0.24 | 0.06 |
| MedGemma-27B-Text | 307 | 190 | 3.09 | 3.15 | 0.22 | 0.03 |
| Baichuan-M2-32B | 332 | 190 | 3.13 | 3.59 | 0.15 | 0.00 |
| HuatuoGPT-o1 | 314 | 190 | 3.07 | 3.63 | 0.21 | -0.01 |

**Framework-based evaluation provides a multidimensional assessment beyond conventional outcome-level metrics**

Conventional outcome-level metrics, such as diagnostic accuracy and medication-level F1, provide important but relatively narrow assessments of model performance. Diagnostic accuracy evaluates whether the final diagnosis matches the reference standard, whereas medication-level F1 primarily measures the overlap between model-generated medications and reference medications. These metrics are useful for assessing final outputs, but they do not fully characterize how a model reaches a diagnosis, whether the reasoning process is clinically coherent, or whether medication recommendations contain safety-relevant errors. To examine how the proposed framework-based scores differ from these conventional outcome-level metrics, we compared Dx scores with diagnostic accuracy and MS scores with medication-level outcome measures. Details of the outcome-level computations are provided in **Supplementary Note 1**.

As the conventional outcome-level counterpart to the framework-based Dx score, diagnostic accuracy produced a different and more compressed ranking pattern across models (**Fig. 5**). Claude 4.6 Sonnet achieved the highest diagnostic accuracy, with a score of 0.624, followed closely by GPT-5 (0.615), GPT-5.5 (0.611), Baichuan-M2 (0.609), and DeepSeek-V4-Pro (0.606). These five models formed a top-performing cluster, with only small differences in final-diagnosis accuracy. A more pronounced decline was observed for Qwen3.5 (0.568), followed by MedGemma (0.518) and Claude 4.0 Sonnet (0.511). Diagnostic accuracy was below 0.50 for the remaining models, including HuatuoGPT-o1 (0.495), DeepSeek-V3 (0.475), Qwen3 (0.471), and GPT-4o (0.462). Notably, this ranking was not fully aligned with Dx-based performance; GPT-4o, for example, had a Dx score comparable to those of DeepSeek-V3 and Claude 4.0 Sonnet, but the lowest diagnostic accuracy.

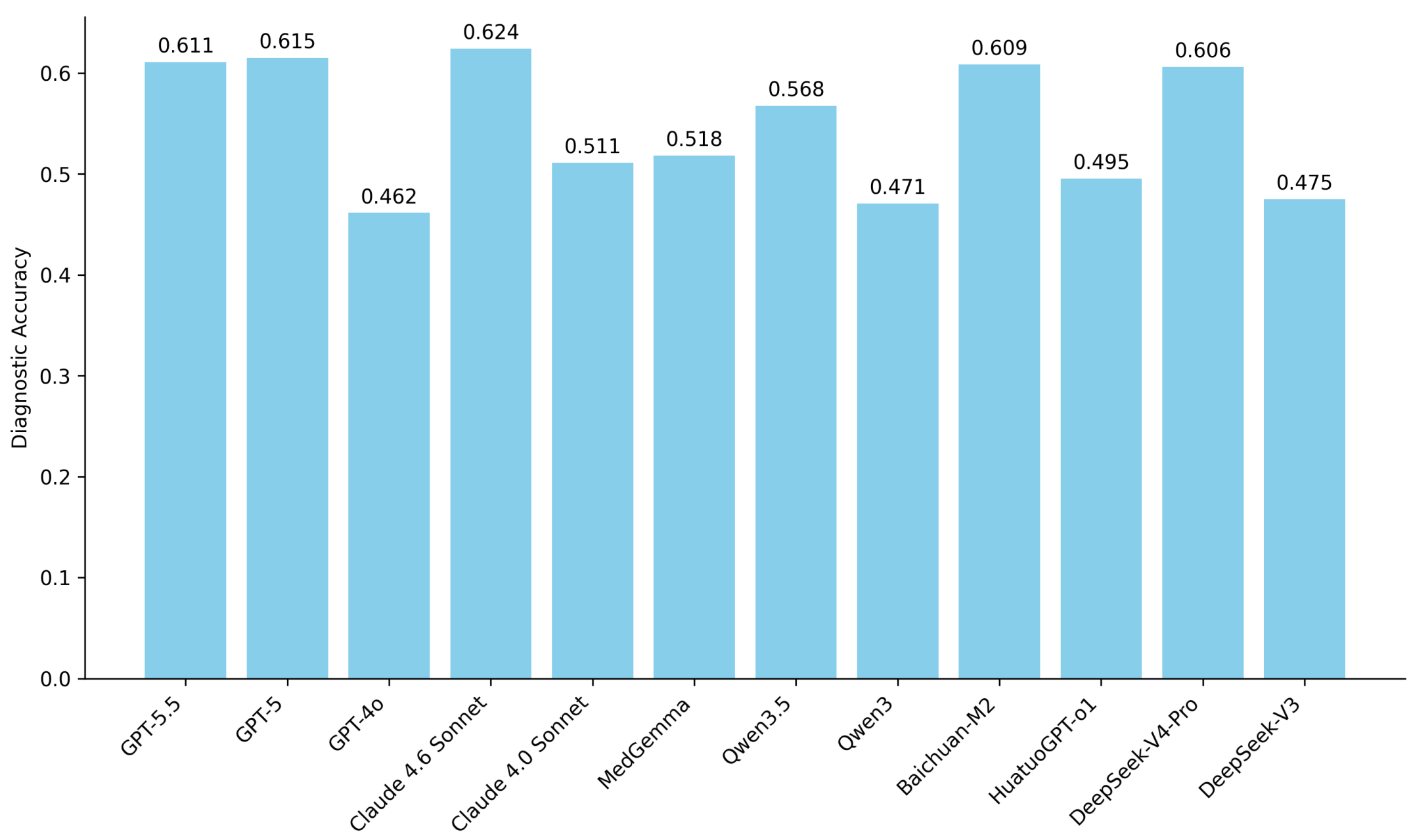

**Fig. 5 | LLM performance in outcome-level diagnostic accuracy.** Claude 4.6 Sonnet achieves the highest diagnostic accuracy.

We further evaluated medication safety by calculating medication-level F1 scores, unsafe drug use rates, and drug–diagnosis mismatch failure rates from MS counts (**Fig. 6**). Average medication-F1 scores ranged from 0.278 (GPT-4o) to 0.341 (Qwen3.5), with Claude 4.0 Sonnet, Claude 4.6 Sonnet and MedGemma also achieving relatively high F1 values. However, higher F1 scores did not consistently correspond to safer medication use. For example, MedGemma achieved a comparatively high (the second place among all models) medication-F1 score (0.327 ± 0.013) but exhibited the lowest medication safety rate (0.611). In contrast, GPT-5, despite a lower F1 score (0.291 ± 0.012), showed relatively high medication safety rate (0.907) and drug–diagnosis mismatch rates (0.067 ± 0.011).

Together, these analyses show that conventional outcome-level metrics and framework-based scores assess overlapping but distinct aspects of clinical model performance. Diagnostic accuracy and medication-level F1 quantify whether final outputs match reference answers, but they do not determine whether a diagnosis was reached through coherent reasoning or whether a medication plan is clinically safe and diagnosis-concordant. Dx and MS address these gaps by evaluating intermediate reasoning, diagnostic coherence, medication safety, diagnosis–treatment concordance and the clinical reasoning supporting drug use. The framework therefore identifies clinically relevant failure modes that may be obscured by final-output metrics alone.

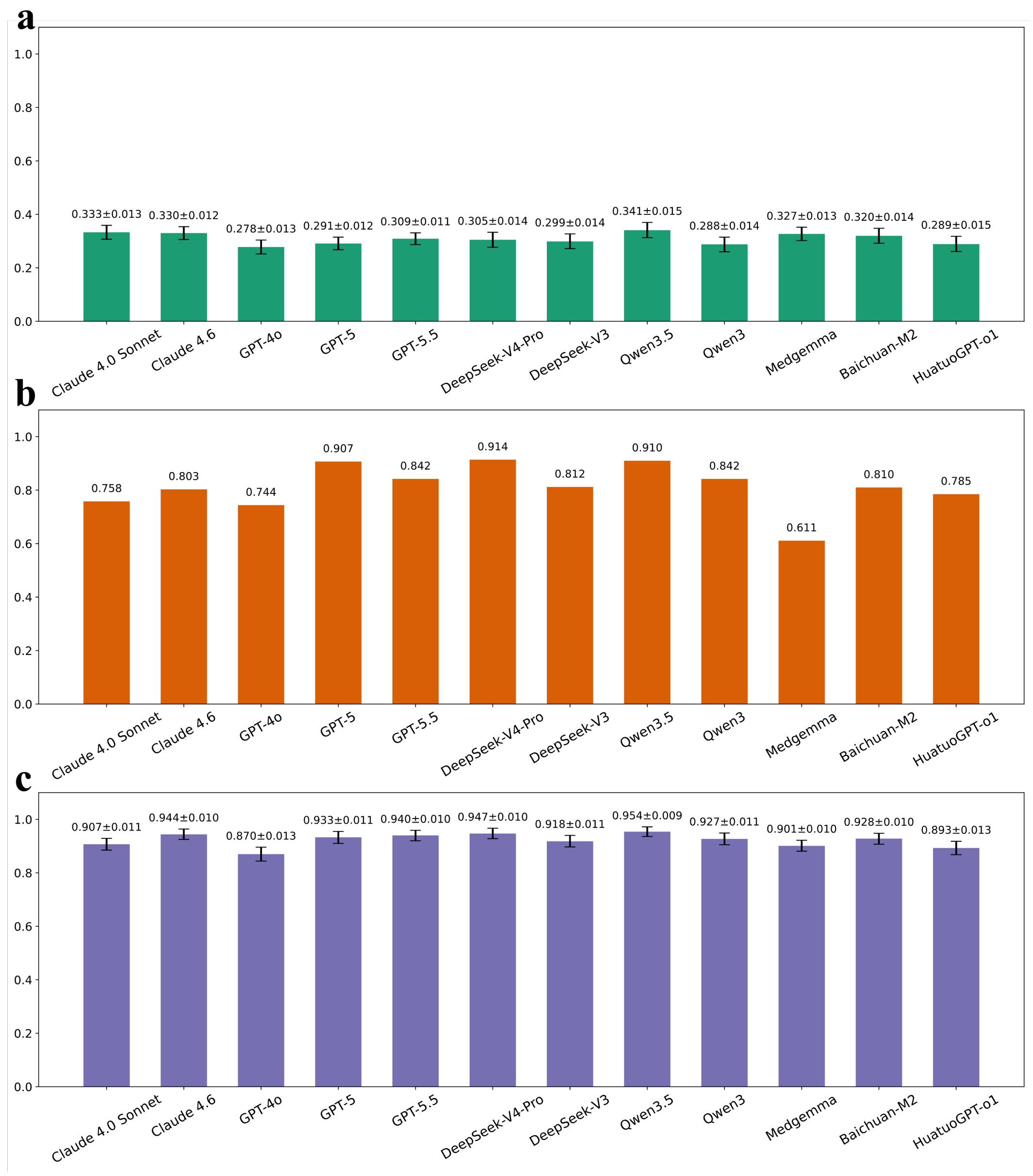

**Fig. 6 | LLM performance in outcome-level medication safety accuracy across all evaluated cases. a,** Medication-F1 Score. **b,** Medication Safety (1 - Unsafe Rate). **c,** Drug-Diagnosis Agreement (1 - Mismatch Rate). Qwen3.5 performed best on Medication-F1 and drug–diagnosis agreement, whereas Deepseek-V4-Pro achieved the lowest unsafe rate.

#### External validation across out-of-distribution datasets

**Table 5**), and compared these results with those obtained on the primary CCQA dataset (**Table 1**). Because Claude 4.0 Sonnet was no longer accessible at the time of external evaluation, the external validation analyses were conducted on the remaining 11 models. Across all three datasets, the overall ranking of model performance remained largely stable. GPT-5.5 achieved the highest weighted average

score on each dataset, with scores of 3.76 on CCQA, 3.71 on PMC-Patients, and 3.75 on PCI. GPT-5 also consistently ranked among the top-performing models, with weighted average scores of 3.69, 3.66, and 3.68, respectively. Models with relatively lower scores on CCQA, such as Qwen3, MedGemma, and GPT-4o, generally remained within the lower-to-middle performance range on the external datasets. In contrast, intermediate-performing models, including DeepSeek-V4-Pro, Qwen3.5, and HuatuoGPT-o1, exhibited modest dataset-dependent rank fluctuations. Overall, these findings suggest that the framework captures reproducible model-level differences in clinical response capability rather than being driven primarily by dataset-specific artifacts.

Across individual evaluation dimensions, models demonstrated broadly consistent performance patterns across datasets. Scores for QS remained among the highest across datasets, ranging from 4.43 to 4.99 on CCQA, 4.49 to 4.97 on PMC-Patients, and 4.63 to 4.98 on PCI. ET and EX also generally received high scores across datasets, although ET showed greater variability on PCI. In contrast, models consistently scored lower on IC, HR, and Dx. For example, Dx scores remained within a relatively narrow and low range across all datasets, spanning 2.63 to 3.55 on CCQA, 2.48 to 3.49 on PMC-Patients, and 2.40 to 3.54 on PCI. These consistent dimension-level trends indicate that the evaluation framework identifies similar strengths and weaknesses in model-generated clinical responses across clinically related datasets.

**Table 4 | Performance of LLMs across the eight evaluation dimensions on PMC-Patients Dataset.** GPT-5.5 achieved the highest overall score (3.71), with different models attaining the top score in individual dimensions**.**

| Model | Size | QS | IC | HR | ET | EX | II | Dx | MS | Wtd. Avg. |
|---|---|---|---|---|---|---|---|---|---|---|
| Claude 4.0 Sonnet | - | - | - | - | - | - | - | - | - | - |
| Claude 4.6 Sonnet | - | 4.82 | 2.92 | 2.68 | 4.65 | 4.26 | 3.64 | 2.74 | 3.56 | 3.46 |
| GPT-4o | - | 4.94 | 2.79 | 2.53 | 4.58 | 4.14 | 3.88 | 2.86 | 3.39 | 3.45 |
| GPT-5 | - | **4.97** | 2.66 | 2.98 | 4.55 | 4.50 | 4.13 | 3.29 | 3.45 | 3.66 |
| GPT-5.5 | - | **4.97** | 2.58 | 3.23 | 4.59 | 4.16 | 4.19 | **3.49** | 3.48 | **3.71** |
| Deepseek-V3 | 671B | 4.89 | 2.39 | 2.71 | 4.50 | 4.53 | 4.11 | 2.80 | 3.55 | 3.50 |
| Deepseek-V4-Pro | 1.6T | 4.90 | **2.95** | 2.82 | 4.68 | 4.51 | 3.79 | 2.70 | **3.70** | 3.55 |
| Qwen3 | 235B | 4.49 | 2.40 | 2.85 | 4.60 | 3.66 | 3.48 | 2.48 | 3.64 | 3.27 |
| Qwen3.5 | 397B | **4.97** | 2.74 | 2.68 | **4.93** | 4.54 | 4.01 | 2.69 | 3.55 | 3.52 |
| MedGemma | 27B | 4.79 | 2.23 | 2.55 | 4.78 | 4.26 | 3.82 | 2.67 | 3.40 | 3.35 |
| Baichuan-M2 | 32B | 4.84 | 2.79 | 2.47 | 4.55 | 4.50 | 3.71 | 2.79 | 3.58 | 3.47 |
| HuatuoGPT-o1 | 72B | 4.53 | 2.82 | **3.37** | 4.52 | **4.56** | **4.21** | 2.73 | 3.65 | 3.59 |

**Table 5 | Performance of LLMs across the eight evaluation dimensions on PCI dataset.** GPT-5.5 achieved the highest overall score (3.75), with different models attaining the top score in individual dimensions**.**

| Model | Size | QS | IC | HR | ET | EX | II | Dx | MS | Wtd. Avg. |
|---|---|---|---|---|---|---|---|---|---|---|
| Claude 4.0 Sonnet | - | - | - | - | - | - | - | - | - | - |
| Claude 4.6 Sonnet | - | 4.87 | 2.36 | 2.64 | 4.67 | 4.46 | 3.34 | 3.04 | 4.01 | 3.56 |
| GPT-4o | - | 4.87 | **2.91** | 2.62 | 3.55 | 4.46 | 3.49 | 2.78 | 3.95 | 3.42 |
| GPT-5 | - | **4.98** | 1.63 | 3.60 | 3.75 | **4.93** | 4.01 | 3.16 | 3.95 | 3.68 |
| GPT-5.5 | - | 4.95 | 1.44 | **3.73** | 4.01 | 4.78 | **4.09** | **3.54** | 3.85 | **3.75** |
| Deepseek-V3 | 671B | 4.90 | 1.81 | 2.90 | 3.72 | 4.76 | 3.49 | 2.40 | 4.38 | 3.45 |
| Deepseek-V4-Pro | 1.6T | 4.92 | 2.40 | 3.28 | 4.28 | 4.78 | 3.55 | 2.72 | **4.56** | 3.70 |
| Qwen3 | 235B | 4.63 | 2.33 | 3.21 | 3.57 | 4.46 | 3.07 | 2.54 | 4.39 | 3.46 |
| Qwen3.5 | 397B | 4.96 | 2.35 | 3.15 | **4.81** | 4.85 | 3.94 | 2.70 | 4.35 | 3.71 |
| MedGemma | 27B | 4.90 | 2.05 | 2.63 | 4.01 | 4.77 | 3.83 | 2.47 | 4.12 | 3.46 |
| Baichuan-M2 | 32B | 4.89 | 2.09 | 2.61 | 3.66 | 4.77 | 3.28 | 2.50 | 4.45 | 3.46 |
| HuatuoGPT-o1 | 72B | 4.68 | 2.01 | 3.54 | 3.57 | 4.78 | 3.87 | 2.50 | **4.52** | 3.60 |

Comparison across datasets further revealed that performance patterns on PMC-Patients closely resembled those observed on CCQA, consistent with the substantial similarity in case-report structure and general clinical consultation characteristics between the two datasets. In contrast, the PCI dataset exhibited a distinct dimension-level profile, most notably characterized by lower IC scores and higher MS scores across most models. IC scores were broadly comparable between CCQA and PMC-Patients, ranging from 2.08 to 3.02 and 2.23 to 2.95, respectively, but shifted downward on PCI, where scores ranged from 1.44 to 2.91. Conversely, MS scores increased from 3.13 to 3.78 on CCQA and 3.39 to 3.70 on PMC-Patients to 3.85 to 4.56 on PCI. Given that PCI was derived from psychological counseling inpatient records rather than general consultation-style clinical cases, these shifts likely reflect differences in specialty-specific reasoning requirements, patient populations, and response-generation contexts.

Further analyses across individual models supported the same overall trend (**Supplementary Fig. 4**). Model-specific capability profiles remained highly consistent between CCQA and PMC-Patients, whereas evaluations on PCI revealed systematic shifts across multiple dimensions. Importantly, these shifts were broadly shared across models rather than occurring randomly. For instance, most models demonstrated reduced IC performance on PCI relative to CCQA and PMC-Patients, while MS scores increased consistently across nearly all models. Collectively, these findings suggest that the proposed framework provides stable relative evaluations across clinical specialties, record structures, and patient populations.

## Discussion

In this study, we introduced AIPatient Arena, an EHR-grounded framework for evaluating large language models in simulated end-to-end clinical consultations. By integrating structured and unstructured electronic health record data into patient-specific knowledge graphs and assessing model behavior across eight clinically grounded dimensions, AIPatient Arena was designed to evaluate more than whether a model can generate plausible clinical outputs. Instead, it examines whether a model can participate appropriately in the sequential, uncertain, and communicative workflow of clinical consultation, including information gathering, clarification, reasoning, explanation, and treatment-related judgment. This design reflects the fact that real-world consultation quality depends not only on final diagnostic or therapeutic outputs, but also on how models acquire information, respond to ambiguity, integrate prior context, and justify clinical decisions throughout the interaction.

A central finding of our evaluation is that model performance cannot be adequately characterized by a single aggregate score. Across LLMs, AIPatient Arena revealed substantial heterogeneity in capability profiles. GPT-5.5 achieved the strongest overall performance and demonstrated a measurable advantage in Dx, with a score of 3.55, while also attaining an almost ceiling-level score in QS. However, strong aggregate performance did not imply uniform strength across all clinical dimensions. Conversely, models with lower overall scores sometimes showed specific areas of relative strength. DeepSeek-V3 achieved the highest score in EX, Qwen3 demonstrated comparatively strong MS performance, and Claude 4.0 Sonnet was the only model to surpass a score of 3 in IC. These findings indicate that similar or lower aggregate scores may conceal qualitatively different patterns of clinical competence and failure, underscoring the importance of evaluating consultation-oriented LLMs as multidimensional workflow participants rather than as single-score systems.

Despite these heterogeneous capability profiles, a consistent cross-model pattern emerged: models generally performed better on surface-level conversational organization than on clinically consequential process dimensions. Many systems were able to maintain professional tone, generate fluent responses, and follow an apparently coherent questioning structure. However, performance was substantially weaker in dimensions requiring comprehensive information coverage, ambiguity management, diagnostic reasoning, and treatment-related judgment. The highest scores for HR and IC across all LLMs reached only 3.32 and 3.02, respectively. Comparatively stronger Dx performance was observed mainly within the GPT-5 series, with GPT-5.5 scoring 3.55 and GPT-5 scoring 3.37. Most models demonstrated intermediate-level performance in II and MS, with scores generally falling between 3 and 4. Together, these results suggest that current LLMs may reproduce the outward structure of clinical consultations while still exhibiting important limitations in information synthesis, ambiguity resolution, and clinically reliable decision-making.

These weaknesses are broadly consistent with recent efforts to move medical LLM evaluation beyond static examination-style benchmarks toward more clinically realistic assessment settings. Prior studies such as CRAFT-MD[11] have shown that model performance declines when systems move from vignette-based question answering to interactive doctor–patient conversations, while MedHELM[6] emphasized that broad medical competence cannot be adequately characterized through exam-style evaluation alone. Consistent with these observations, the clinically meaningful failures identified in our study were often not overtly nonsensical or obviously implausible outputs. Instead, they appeared as subtle breakdowns in consultation process management. Recurrent issues included repetitive questioning, failure to address uncertain answers, omission of past medical history, and lack of empathy. Among these, failure to address uncertain answers emerged as a particularly pervasive issue, with failure rates exceeding 90% across nearly all LLMs in the dialogue-level analysis. The lack of model certainty self-awareness further calls for the need of external supervision and validation of LLM generated medical information in clinical use. These findings suggest that clinically important consultation failures often arise from weaknesses in clarification, contextual integration, uncertainty management, and patient-centered communication rather than from final-answer inaccuracy alone.

This pattern has direct implications for clinical deployment. The observed strengths and weaknesses suggest that current available LLMs may provide practical value in constrained or supervised workflows, but should not be interpreted as ready for autonomous clinical consultation. Models may be useful for structured history-taking, preliminary information organization, documentation support, patient-facing explanation under clinician review, or assisting clinicians in summarizing consultation content. In these settings, human oversight or human-in-the-loop validation remains central, and model outputs can be checked, corrected, and contextualized by trained professionals. In contrast, tasks requiring ambiguity resolution, longitudinal information synthesis, diagnostic prioritization, medication-related judgment, or independent clinical decision-making appear substantially more vulnerable to failure. Deployment readiness should therefore be assessed at the level of specific workflow components and supervision requirements, rather than treated as a unified property of a model.

This interpretation is consistent with prior work in digital medicine arguing that evaluation of healthcare AI should account for the interactional context in which clinicians and patients encounter the system, rather than treating performance as a decontextualized property of the model alone. Our findings reinforce this point by showing that consultation quality depends on the model's ability to manage dialogue, uncertainty, and explanation over time. In this respect, AIPatient Arena may serve as a pre-deployment stress-testing framework: it allows developers and health systems to examine whether a model's failure profile is compatible with a proposed clinical role, what level of oversight would be needed, and which parts of the workflow remain too fragile for safe use.

To further characterize the sources of these consultation failures, we conducted a dialogue-sufficiency analysis examining the relationship between upstream information collection and downstream reasoning performance. Under dialogue-sufficient conditions, all LLMs demonstrated improvements in Dx scores, with gains ranging from 0.15 to 0.28. This finding suggests that part of the observed weakness in Dx may be attributable to incomplete or insufficient information acquisition during earlier stages of consultation. However, MS scores remained largely stable under the same conditions, indicating that improved diagnostic accuracy and reasoning does not necessarily translate into safer or more appropriate treatment recommendations. Medication-related performance may depend on additional capabilities beyond diagnostic inference, including contraindication recognition, risk assessment, treatment prioritization, and justification of therapeutic choices. These results suggest that consultation failure cannot be attributed to a single capability deficit, but instead emerges from partially independent weaknesses across information gathering, reasoning, and treatment planning.

Building on this broader movement, our results further support the need for rubric-based, safety-oriented evaluation in medical AI assessment. HealthBench[7] argues that healthcare evaluation should prioritize safety, clarity, appropriate escalation, and contextual judgment using physician-written rubrics rather than relying on answer matching alone. More recent clinician-facing evaluation efforts, including HealthBench Professional, likewise emphasize realistic professional tasks, safety, and rubric-based grading in contexts closer to actual practice. AIPatient Arena extends this direction to simulated clinical consultation by grounding interactions in patient-specific EHR-derived knowledge graphs and evaluating model behavior across sequential stages of the consultation.

Our comparative analyses with conventional outcome-oriented metrics further suggest that endpoint correctness alone may provide an incomplete or even misleading picture of clinical consultation quality. In several cases, models with similar Dx scores demonstrated markedly different diagnostic accuracies, while medication-level F1 metrics captured only a limited aspect of treatment planning behavior. For example, GPT-4o achieved Dx performance comparable to that of DeepSeek-V3 and Claude 4.0 Sonnet, yet demonstrated the lowest diagnostic accuracy among LLMs. This discrepancy is not necessarily contradictory, because Dx evaluates the quality, coherence, and clinical plausibility of the reasoning process, whereas diagnostic accuracy measures agreement with a final reference diagnosis. Conversely, GPT-5 showed relatively low medication-level F1 performance despite comparatively strong performance in MS within consultation contexts. These discrepancies indicate that conventional endpoint-oriented metrics may fail to capture important differences in reasoning quality, safety alignment, clarification behavior, and internal decision consistency.

The external validation analyses further support the robustness and contextual sensitivity of the proposed framework. Across PMC-Patients and PCI, relative model rankings remained broadly stable despite differences in dataset composition and clinical context, suggesting that AIPatient Arena captures reproducible differences in consultation capability rather than dataset-specific artifacts. At the same time, the framework remained sensitive to clinically meaningful shifts in consultation characteristics. In particular, PCI dataset, derived from psychological counseling inpatient records, consistently produced lower information coverage scores and higher medication safety and justification scores across models relative to CCQA and PMC-Patients. Importantly, these shifts were broadly shared across models rather than occurring randomly, indicating that the framework responds coherently to differences in specialty-specific reasoning demands, patient populations, and consultation structure. Together, these findings suggest that AIPatient Arena maintains stable comparative evaluation across clinically related datasets while preserving sensitivity to contextual variation in real-world consultation settings.

In consultation settings, multidimensional evaluation is not simply a richer descriptive layer on top of accuracy; it is necessary to avoid overestimating real-world readiness. A model that is directionally correct at the endpoint but fragile in ambiguity resolution, contextual integration, or treatment justification may still be unsafe in practice. Clinically meaningful failures may therefore emerge even when final outputs appear plausible or partially correct. Multidimensional process-oriented evaluation is necessary not only to characterize model behavior more precisely, but also to identify which parts of the consultation workflow are insufficiently reliable for clinical use.

Several limitations should be acknowledged. First, although the framework is grounded in real EHR-derived patient information, the consultations remain simulated ones rather than prospectively observed in real-world live clinical practice. This was a deliberate design choice for safety and scalability, and it follows the logic of prior work such as CRAFT-MD, which also uses simulated interactions to study pre-deployment readiness. However, simulated evaluation cannot fully capture all aspects of real consultations, including patient emotion, clinician adaptation, institutional context, and the downstream consequences of errors over time. Second, our current cohort is limited in scale and source, and further work is needed to examine generalizability across institutions, specialties, clinical cultures, and patient populations. Third, although human review supported the interpretability of the multidimensional framework, future work should strengthen this component with larger clinician panels, formal agreement analyses, and prospective validation in real workflow settings.

An immediate next step is to expand the framework to support broader subgroup and fairness analyses, offering potential of improved equity and generalizability. Another future direction is to use AIPatient Arena longitudinally, for example, as a regression-testing and audit framework when model versions change, prompts are updated, or institution-specific deployments are adapted over time. Post-deployment monitoring, performance decay, feedback loops, and longitudinal evaluation will also be potential domains that our framework can support with further development.

In summary, AIPatient Arena advances the evaluation of medical LLMs by moving from static benchmarking toward workflow-oriented, EHR-grounded, multidimensional assessment of consultation quality. Our findings show that current LLMs may appear conversationally competent while still exhibiting weaknesses that limit safe real-world consultation use, particularly in ambiguity handling, explanation, diagnostic reasoning, and medication safety. More broadly, the study argues that real-world readiness should be treated as a property of the entire consultation process, not inferred from endpoint performance alone. We therefore view AIPatient Arena as both an evaluation framework and a practical tool for pre-deployment assessment of whether LLMs are ready to participate in patient-facing clinical workflows.

## Methods

### Study design and evaluation objective

We developed AIPatient Arena, an EHR-grounded evaluation framework for assessing LLMs in simulated end-to-end clinical consultations. The objective of the framework was not only to measure whether a model could produce a plausible final diagnosis or treatment recommendation, but also to evaluate whether it could participate appropriately in the sequential workflow of a clinical encounter, including questioning, information gathering, ambiguity resolution, reasoning, explanation, and medication-related justification. This design reflects recent calls for medical AI evaluation frameworks that move beyond examination-style accuracy toward clinically realistic, workflow-oriented, and safety-relevant assessment.

AIPatient Arena was designed as a pre-deployment evaluation environment rather than a live clinical system. Simulated consultation was used to enable reproducible, scalable, and safety-conscious testing of LLM behavior before real-world deployment. Different to prior work such as CRAFT-MD[11], which uses non-real-world record-based simulated doctor–patient dialogue to assess conversational clinical performance, our framework leverages EHR-grounded patient construction and multidimensional process-level consultation assessment.

### Clinical consultation dataset and cohort

We constructed our CCQA dataset based on the publicly available DrugEHRQA dataset[26]. DrugEHRQA selected 475 patient records from the MIMIC-III database[13], which contains de-identified EHRs of over 40,000 ICU patients admitted to Beth Israel Deaconess Medical Center between 2001 and 2012. For each selected record, medication information was extracted from discharge summaries to generate patient-specific prescription data. Access to the source data was obtained through the PhysioNet credentialed access program.

The dataset was selected to support consultation-oriented evaluation rather than isolated single-turn prediction. Unlike examination-style benchmarks, CCQA setting requires models to progressively collect clinical information, handle incomplete or uncertain responses, and synthesize patient context distributed across multiple clinical records and interaction turns.

To maintain cohort consistency and focus on adult consultation scenarios, patients younger than 18 years were excluded because pediatric cases differ substantially in physiology, treatment strategies, and ethical considerations. The final CCQA dataset consisted of 442 admission records from 437 unique adult patients, including five patients with two separate admissions.

The demographic characteristics of the cohort, including age, sex, and race distribution, are presented in **Table 6**. In the current cohort, the mean age was 64.09 years, with 202 female patients (46.22%) and 235 male patients (53.78%). Additional details are provided in **Supplementary Table 2**.

**Table 6 | Demographic characteristics of the CCQA cohort.**

| Characteristic | Count (N) / Value | Proportion (%) |
|---|---|---|
| **Total patients** | 437 | - |
| **Total admissions** | 442 | - |
| **Age, years (mean ± SD)** | 64.09 ± 17.03 | - |
| **Female (F)** | 202 | 46.22 |
| **Male (M)** | 235 | 53.78 |
| **WHITE** | 325 | 74.37 |
| **BLACK** | 56 | 12.81 |
| **OTHER** | 27 | 6.18 |
| **HISPANIC/LATINO** | 18 | 4.12 |
| **ASIAN** | 11 | 2.52 |

## Construction of the EHR-grounded AIPatient knowledge graph

The AIPatient Knowledge Graph (KG) was constructed by integrating structured patient data with entities extracted from unstructured discharge summaries using an LLM–based named entity recognition (NER) system. Extracted entities included medical history, symptom descriptions, social history, family history, allergies, diagnoses, vital signs, and medication information. The resulting graph served as the patient-specific ground truth for consultation simulation. Summary statistics of the graph are provided in **Table 7**. The final graph contains 9,682 nodes and 19,752 relationships, of which 7,127 nodes (73.61%) correspond to entities extracted from unstructured clinical narratives.

The EHR-grounded KG representation was designed to better reflect the complexity of real-world clinical consultations than purely synthetic or vignette-based benchmarks. Unlike static examination-style cases, real clinical records contain incomplete information, ambiguous descriptions, shorthand expressions, and context distributed across multiple documents and time points. By grounding each simulated consultation in patient-specific structured and narrative EHR data, the framework enables evaluation under more realistic and context-dependent clinical interaction settings.

**Table 7 | AIPatient knowledge graph structure and entity distribution.**

| Statistic category / Entity type | Count | Proportion (%) |
|---|---|---|
| **Total nodes** | 9,682 | 100.0 |
| **Total relationships** | 19,752 | - |
| **Average degree** | 4.08 | - |
| **Medical history** | 2,499 | 25.81 |
| **Symptoms (Primary)** | 1,954 | 20.18 |
| **Social history** | 1,464 | 15.12 |
| **Family history** | 322 | 3.33 |
| **Allergy** | 299 | 3.09 |
| **Other structured entities (e.g., FamilyMember, Duration)** | 1468 | 15.16 |
| **Vital signs** | 734 | 7.58 |
| **Diagnosis** | 479 | 4.95 |
| **Medication profiles** | 463 | 4.78 |

## Simulation of end-to-end clinical consultations

Each LLM was instructed to simulate an experienced clinician conducting a structured outpatient consultation. During the interaction, the model was required to ask one question at a time and proceed in a logical and concise manner. Each consultation was limited to a maximum of 15 interaction turns, although the consultation could terminate earlier if sufficient information had been collected. Upon reaching the final turn, the model was required to provide a diagnosis and treatment plan even if some information remained incomplete. The prompt additionally instructed the model to continuously record all patient information obtained during the consultation and explicitly mark unavailable information as “Not available”.

This consultation protocol was designed to emulate a constrained but realistic clinical workflow in which the model must determine what information to request, when to stop gathering additional details, and how to synthesize the interaction into clinically actionable outputs. Restricting the interaction to one question per turn and a finite consultation horizon enabled evaluation of consultation discipline, information prioritization, and diagnostic efficiency under realistic interaction constraints rather than through unconstrained iterative prompting.

At the end of each consultation, the model generated a structured clinical record including the full conversation history, category summaries of the collected patient information, diagnostic reasoning, the final clinician diagnosis, and diagnosis-based recommended drugs with corresponding drug reasoning. These outputs were consolidated as the final consultation record for each case and subsequently evaluated by the automated assessment framework.

### Evaluated models

We evaluated a diverse set of frontier LLMs, including proprietary and open-source systems, as well as general-purpose and medically adapted models. The LLMs included GPT-5.5, GPT-5[15], GPT-4o[16], Claude 4.6 Sonnet[17], Claude 4.0 Sonnet[18], DeepSeek-V4-Pro[19], DeepSeek-V3[20], Qwen3.5-397B[21], Qwen3-235B[22], MedGemma-27B-Text-IT[23], HuatuoGPT-o1-72B[24], and Baichuan-M2-32B[25]. This model set was selected to compare general conversational capability and medically oriented adaptation within the same consultation-oriented evaluation framework.

All models were evaluated under a standardized zero-shot setting. Neither the LLMs nor the prompts were fine-tuned or optimized using task-specific training data. A unified prompting and consultation protocol was applied across all models to ensure consistency of the evaluation setting. Full prompt details are provided in **Supplementary Note 2**.

### Multidimensional consultation evaluation framework

Consultation quality was evaluated using a multidimensional rubric-based framework comprising eight clinically grounded dimensions, each operationalized through predefined domain-specific failure patterns: medical interview questioning skills, information coverage, handling of ambiguous patient responses, ethical and professional conduct, clarity and transparency of clinical explanations, information integration, diagnostic accuracy and reasoning, and medication safety and justification (**Fig. 7**). To ensure consistent evaluation across all experiments and models, the framework was instantiated using GPT-5 as the evaluator model. The selected dimensions were designed to capture complementary aspects of consultation performance spanning information gathering, clinician–patient communication, reasoning quality, and treatment planning.

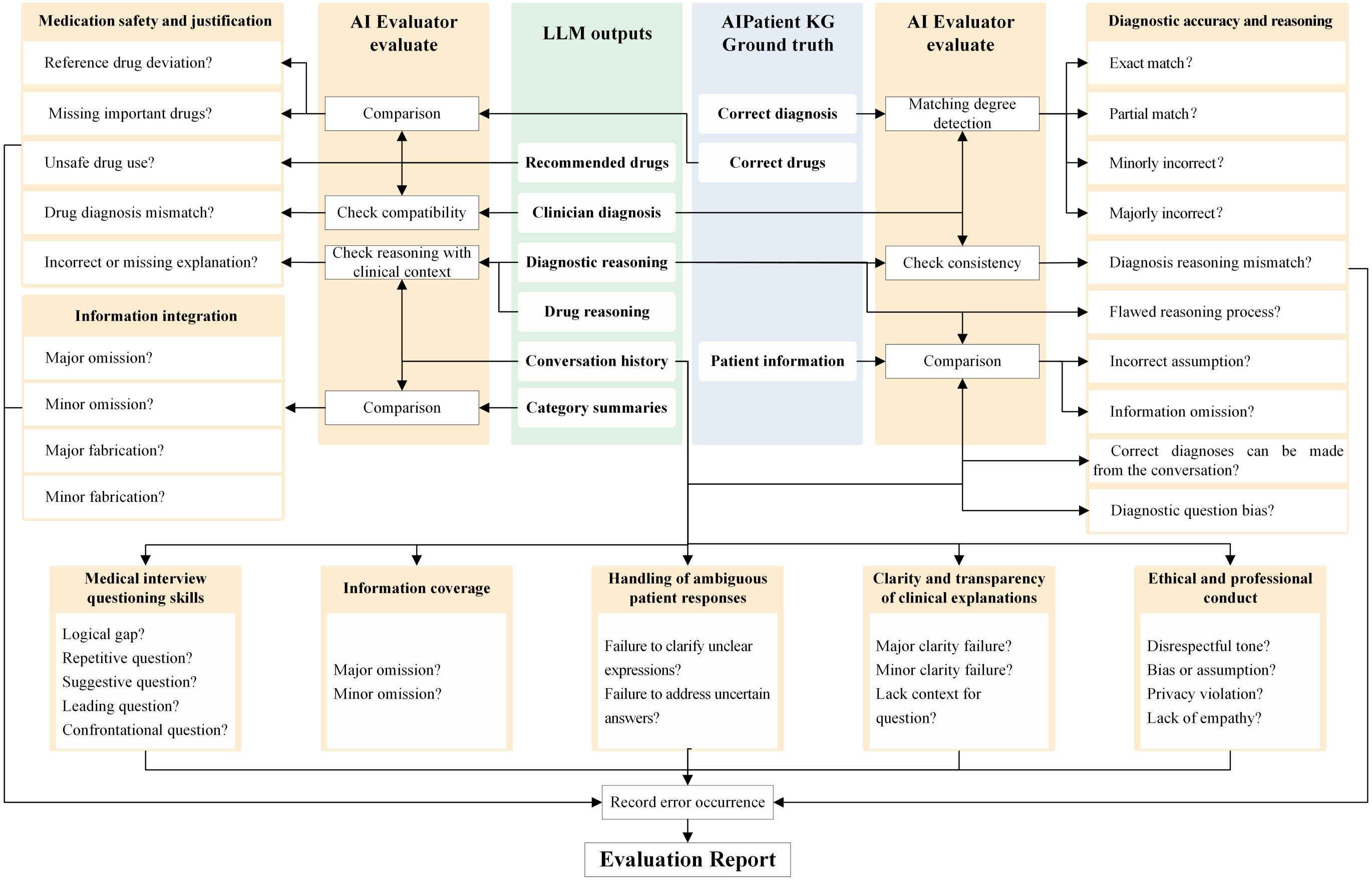

**Fig. 7 | Evaluation process across eight dimensions.** AI evaluator assessed the clinical capabilities of LLMs by jointly analyzing the conversation history between LLMs and AIPatient together with the final clinical outputs generated by LLMs, including category summaries, clinical diagnoses, diagnostic reasoning, recommended medications, and medication reasoning. To ensure objective evaluation, the assessment further incorporated ground-truth information from AIPatient Knowledge Graph, including correct diagnoses, reference medications, and structured patient information. The five lower dimensions, including medical interview questioning skills, handling of ambiguous patient responses, information coverage, clarity and transparency of clinical explanations, and ethical and professional conduct, primarily evaluated communication quality, soft skills, and clinical compliance based on the dialogue history. Medication safety and justification in the upper left evaluated the safety and appropriateness of recommended medications, their consistency with reference medications and diagnoses, and the completeness and correctness of medication reasoning by jointly analyzing conversation history, diagnostic reasoning, and reference medications. Information integration in the lower left cross-validated AI-generated category summaries against the original dialogue and patient ground-truth information to identify information omission and fabrication. Diagnostic accuracy and reasoning on the right evaluated diagnostic accuracy and logical consistency against the reference diagnosis, verified reasoning fidelity with respect to the conversation history and patient information, and analyzed potential sources of diagnostic error, including insufficient diagnostic support and question-induced diagnostic bias.

AI evaluator assessed the clinical capabilities of LLMs by jointly analyzing three sources of information: (1) the complete conversation history between LLM and AIPatient; (2) the final clinical outputs generated by LLMs, including category summaries, clinical diagnoses, diagnostic reasoning, recommended medications, and medication reasoning; and (3) ground-truth information from AIPatient Knowledge Graph, including reference diagnoses, reference medications, and structured patient information. Different evaluation dimensions incorporated different combinations of these information sources depending on the clinical competency being assessed.

Among the eight evaluation dimensions, five focused primarily on interactional and communication performance based on the conversation history. For medical interview questioning skills, AI evaluator assessed whether LLMs asked focused, logically coherent, and clinically relevant questions while avoiding repetitive, suggestive, leading, or confrontational questioning patterns. For information coverage, AI evaluator evaluated the completeness of information gathering by identifying both major and minor omissions of clinically relevant details during the consultation. For handling of ambiguous patient responses, AI evaluator examined whether vague symptom descriptions or uncertain patient statements were appropriately clarified within a reasonable number of conversational turns. For clarity and transparency of clinical explanations, AI evaluator assessed whether LLMs explained clinical information in a manner that was understandable to non-expert patients, including the appropriate use of accessible language and avoidance of unnecessarily complex medical terminology. For ethical and professional conduct, AI evaluator examined whether the consultation contained inappropriate clinician behaviors, including disrespectful tone, unsupported assumptions or bias, privacy violations, and lack of empathy during patient communication.

The remaining three dimensions focused on higher-order cognitive and clinical reasoning performance beyond the consultation dialogue itself. For clinical information integration, AI evaluator cross-validated the category summaries generated by LLMs against the original conversation history to identify omissions and fabricated content, which were further categorized as either major or minor according to clinical severity. For diagnostic accuracy and reasoning, AI evaluator assessed multiple aspects of diagnostic performance, including: (1) alignment between the clinician diagnosis generated by LLMs and the correct diagnosis from AIPatient Knowledge Graph; (2) validity, coherence, and internal logical consistency of the diagnostic reasoning process, including consistency between cited evidence and the conversation history as well as compatibility between the reasoning process and the final clinician diagnosis; and (3) sufficiency and clinical relevance of the diagnostic information collected during the consultation, including whether the conversation provided adequate support for the correct diagnosis and whether diagnostic question bias was present. For medication safety and justification, AI evaluator assessed the appropriateness and clinical rationale of recommended drugs relative to the correct drugs and diagnoses, including omission of clinically important drugs, drug–diagnosis inconsistency, unsafe drug use, and whether the drug reasoning was adequately supported by the clinical context and diagnostic reasoning process.

Scoring was performed using a predefined penalty-based scheme. For each evaluation dimension, we specified a set of domain-specific failure patterns, the deduction assigned to each failure pattern, and a maximum allowable penalty for that dimension (**Supplementary Table 3**). The AI evaluator first identified the presence of these predefined failure patterns in the relevant input sources and then applied the corresponding deductions to compute the dimension-level score. Penalties were accumulated within each dimension until the maximum allowable deduction was reached, preventing any single dimension from contributing negative scores or disproportionate penalties. Compared with conventional Likert-scale evaluation, this error-focused design enabled more fine-grained assessment of both the frequency and clinical severity of consultation failures. An example evaluation report containing dimension-level scores and step-level failure pattern annotations is provided in **Box 1**.

To derive an overall consultation performance score, dimension-level scores were aggregated using a predefined weighted average. Diagnostic reasoning and medication safety and justification were assigned the highest weights (0.25 and 0.20, respectively) because of their direct relevance to clinical decision-making and patient safety. Medical interview questioning skills, information coverage, handling of ambiguous patient responses, clarity and transparency of clinical explanations, and information integration were each assigned a weight of 0.10. These dimensions were considered essential components of a high-quality consultation, but they primarily support, rather than replace, the core clinical tasks of diagnostic reasoning and safe treatment recommendation. Ethical and professional conduct was assigned a weight of 0.05 because it was treated as a baseline professional requirement expected across all consultations; the lower numerical weight does not imply lower importance, but reflects its role as a minimum standard rather than the main discriminator of clinical decision-making performance. All weights were specified a priori and applied consistently across all LLMs. The weighted average score reported in **Table 3** was calculated as the normalized weighted sum across the eight evaluation dimensions.

Weak pairwise correlations among the evaluation dimensions supported their interpretation as complementary rather than redundant measures of consultation quality. Collectively, the framework operationalized consultation-readiness as a multidimensional construct encompassing communication quality, information gathering, reasoning validity, and treatment safety, thereby enabling systematic comparison between automated LLM evaluation and subsequent expert human assessment.

**Box 1 Evaluation report example**. Example output showing structured multi-dimensional evaluation with step-level feedback and failure pattern annotations across clinical competencies.

SubjectID: 28898, AdmissionID: 115391
"RELIGION": "PROTESTANT QUAKER", "MARITAL_STATUS": "DIVORCED", "GENDER": "M", "ETHNICITY": "WHITE", "AGE": 71.0

**Medical interview questioning skills (Score: 4.5)**
- Step 15 (repetitive questions): Repeats CHF history despite being established earlier.

**Information coverage (Score: 2.0)**
- (missing major): Allergy information not elicited (safety-relevant).
- (missing minor): Family, social history, and mental status not assessed, limiting contextual understanding.

**Handling of ambiguous patient responses (Score: 4.0)**
- Step 2 (failure to clarify unclear expressions): The patient provides ambiguous timing for multiple symptoms, but the doctor does not clarify their mapping to timelines.
- Step 9 (failure to address uncertain answers): The patient gives an uncertain response regarding prior tests, but no follow-up is conducted within subsequent turns.

**Ethical and professional conduct (Score: 3.0)**
- Step 2 (lack of empathy): The patient expresses fear about potential respiratory failure, but the doctor does not acknowledge or validate this concern.
- Step 6 (lack of empathy): The patient reports distress due to worsening chest pain; however, the doctor shifts directly to other symptoms without emotional response.
- Step 14 (lack of empathy): The patient expresses discomfort from dyspnea, but no reassurance or emotional acknowledgment is provided.

**Clarity and transparency of clinical explanations (Score: 3.5)**
- Step 12 (major clarity failure): The term “palpitations” is used without explanation, which may not be understood by a layperson.
- Step 13 (minor clarity failure): The phrase “acute respiratory symptoms” includes potentially unclear terminology (“acute”) without clarification.

**Information integration (Score: 3.4)**
- (major omission): Symptom onset and timing are not clearly summarized.
- (major omission): Chest pain severity and its impact on daily activities are missing.
- (major omission): Associated respiratory symptoms (e.g., cough) are not included.

**Diagnostic accuracy and reasoning (Score: 2.5)**
- (diagnosis accuracy: minorly incorrect): The final diagnosis is clinically plausible but does not match the reference diagnoses.
- (diagnostic questioning bias): Mild diagnostic bias toward CHF, with focus on edema, weight gain, and prior heart failure, and limited exploration of ACS, pulmonary embolism, and pneumonia.
- (correct diagnoses can be made from the conversation): CHF with acute dyspnea and chest pain suggests possible respiratory failure.
- (incorrect assumption): The absence of peripheral edema is used to argue against fluid overload, which is not a reliable clinical inference.
- (information omission): Key diagnostic features (e.g., detailed chest pain characteristics) are not considered.
- (flawed reasoning process): An overreliance on edema and ignores chest pain, leading to a narrow assessment that misses ACS.
- (diagnosis reasoning mismatch): No explicit mismatch between reasoning and final diagnosis is observed, although the reasoning remains incomplete.

**Medication safety and justification (Score: 4.1)**
- (reference drug deviation): Lisinopril and Carvedilol are recommended but not included in the reference set.
- (missing important drugs): Ceftriaxone Missing and not substituted. Counted as error.
- (drug diagnosis mismatch): No mismatch observed; recommended medications are therapeutically aligned with the diagnosis.
- (unsafe drug use): No unsafe drug use identified.
- (incorrect or missing explanation): Drug usage explanations are appropriate and complete.

### Human review of automated assessments

To assess the fidelity of the automated clinical evaluation framework, we conducted a structured human-expert validation in which clinicians rated how well the automated evaluation outputs align with their own clinical judgment. The user study involving medical experts, including attending physicians, was approved by the Institutional Review Board of Qilu Hospital of Shandong University (IRB Protocol Number: KYLL-2025-11-015-1). All participants provided informed consent prior to participation. All procedures involving human participants were conducted in accordance with the Declaration of Helsinki.

The validation was performed using a dedicated set of clinical cases sampled across multiple large language models. The validation set comprised 120 simulated clinical encounters, including 30 cases randomly drawn from the outputs of each of four generative architectures: GPT-4o, Claude 4.0 Sonnet, Qwen 3, and MedGemma. This balanced sampling strategy ensured that human assessments reflected model-agnostic properties of the evaluation framework rather than idiosyncratic characteristics of any single system.

Independent human evaluation was performed by 20 medical experts with formal clinical training and experience in standardized clinical assessment. To minimize potential bias, the evaluators were randomly divided into four groups of five, with each group assigned to assess cases generated by one model.

Human evaluators rated the alignment between the automated evaluation outputs and their own clinical judgment using a standardized five-point Likert scale ranging from 1 (very poor alignment) to 5 (excellent alignment). To obtain stable case-level estimates and reduce inter-rater variability, individual reviewer scores were aggregated by computing the mean score for each case and evaluation dimension. These case-level mean scores constituted the primary quantitative benchmark for subsequent statistical analysis and visualization. The evaluation was administered via a web-based interface, and details of the interface design and annotation workflow are provided in **Supplementary Fig. 5**.

### External validation

To evaluate the robustness and generalizability of AIPatient Arena beyond the primary CCQA evaluation setting, we conducted external validation on two independent cohorts representing different degrees of distribution shift relative to CCQA.

The first external cohort was derived from the publicly available PMC-Patients dataset, which consists of narrative medical case reports extracted from biomedical literature rather than routine EHR documentation. Although this cohort differed from CCQA in data source and writing style, it retained a broad general-medicine focus and therefore shared greater clinical similarity with CCQA in terms of diagnostic reasoning, disease-oriented clinical assessment, and treatment planning. Candidate case reports were first screened using a large language model according to predefined eligibility criteria. Specifically, the model was used to identify and exclude cases lacking key clinical elements required for virtual patient construction, including sufficient medical history, diagnostic assessment, or treatment information. Manual review was then performed to verify the screening results, resolve uncertain cases, and confirm final eligibility. To ensure adequate therapeutic complexity for medication-related evaluation, eligible cases were additionally required to include at least five treatment medications. This criterion was used to ensure that medication safety and medication-reasoning dimensions could be meaningfully evaluated, although it enriched the cohort for therapeutically complex cases. The final PMC-Patients cohort included 119 patients, with a mean age of 48.86 years (SD, 18.28); 54 patients were female (45.38%) and 65 were male (54.62%) (**Supplementary Table 4**).

The second external validation cohort, termed the PCI cohort, consisted of real-world inpatient clinical records collected from a Chinese hospital between July 2024 and November 2024. The cohort was predominantly composed of adolescents, although a small number of adult patients were also included. This cohort represented a more pronounced distribution shift from CCQA because it differed simultaneously in language, patient age distribution, specialty domain, care setting, disease spectrum, and documentation style. The study was approved by the institutional ethics committee, and all records were de-identified before analysis. Direct identifiers, including patient names, were removed, and date-related information was transformed to reduce re-identification risk. Records were manually screened, and those lacking key clinical elements required for virtual patient construction, including sufficient medical history, diagnostic assessment, or treatment information, were excluded. After screening and quality control, 67 patient records met the inclusion criteria and were included for external evaluation. The final cohort had a mean age of 19.30 years (SD, 12.21); 42 patients were female (62.69%) and 25 were male (37.31%) (**Supplementary Table 5**). Because the original records were written in Chinese, they were translated into English before virtual patient construction and model evaluation to maintain consistency with the primary experimental protocol. Translated cases were reviewed to ensure preservation of clinically relevant information.

For both external cohorts, we applied the same virtual patient construction framework used for the CCQA cohort. Clinical records were transformed into standardized virtual patient cases using the same case organization strategy, patient knowledge representation, and interaction protocol. Dataset-specific preprocessing was limited to format normalization, de-identification where applicable, translation for the Chinese cohort, and quality control required for virtual patient construction. No dataset-specific model retraining, fine-tuning, prompt optimization, scoring-rule modification, or hyperparameter adjustment was performed before external evaluation.

The same evaluation framework was then used to assess the performance of the evaluated LLMs on the external virtual patient databases. Prompting strategies, consultation procedures, scoring rubrics, scoring methods, and evaluation criteria were kept consistent with those used in the CCQA experiments to ensure comparability across cohorts. External validation results were compared with those from the primary CCQA cohort at multiple levels, including overall performance scores, scores across the eight predefined evaluation dimensions, relative model rankings, and dimension-level performance changes. These analyses were used to assess model stability and framework robustness across different degrees of distribution shift.

**Statistical analysis**

Analyses were performed at the case or consultation level. Human Likert-scale ratings were summarized as medians and IQRs, while automated evaluation scores were summarized as mean values across cases. Pairwise associations between evaluation dimensions were assessed using Pearson correlation coefficients based on pooled case-level scores. Model performance was reported descriptively using dimension-level scores, weighted average scores, absolute score differences, and relative rankings. Analyses were descriptive and were not designed for formal hypothesis testing. All analyses were conducted in Python using standard scientific computing libraries.

**Code availability**. The most up-to-date version is available on GitHub: https://github.com/PAI-CUHK/AIPatientArena.

**Data availability**. The datasets generated and/or analyzed during the current study are not publicly available due to institutional policy and human subject privacy protection requirements but are available from the corresponding author on reasonable request. Our research was based on the MIMIC-III dataset accessed through PhysioNet[13]. Prior training of CITI Data or Specimens Only Research (Record Number:

59460661). The demographics and recording metadata are included in **Table 6**, **Supplementary Table 2, 4 and 5** for the CCQA, PMC-Patients and PCI cohorts, respectively.

**Acknowledgements.** This study is supported in part by the National Key Research and Development Program of China under Grant 2023YFB4706100, in part by the National Natural Science Foundation of China under Grant 62573271, the Major Basic Research Project of Shandong Provincial Natural Science Foundation under Grant ZR2025ZD24, and the Chinese University of Hong Kong Vice-Chancellor Early Career Professorship Scheme under Grant 441.

## Author information

### Contributions

**J.N.:** Writing - review and editing, Writing - original draft, Validation, Software, Resources, Methodology, Investigation, Data curation, Formal analysis, Conceptualization. **H.Y.:** Writing - review and editing, Writing - original draft, Validation, Software, Resources, Methodology, Investigation, Formal analysis, Data curation. **W.W.:** Writing - review and editing, Writing - original draft, Validation, Software, Resources, Methodology, Investigation. **G.D.:** Writing - review and editing, Writing - original draft, Data Curation. **J.H.:** Writing - review and editing, Writing - original draft, Data Curation. **X.Li.:** Writing - review and editing, Writing - original draft, Data Curation. **Z.L.:** Writing - review and editing, Writing - original draft, Formal analysis. **X.Lin.**: Writing - review and editing, Writing - original draft, Data Curation. **K.C.S.**: Writing - review and editing, Writing - original draft, Resources. **B.Y.Y.**: Writing - review and editing, Writing - original draft, Resources. **Y.K.W:** Writing - review and editing, Writing - original draft, Resources, Funding acquisition. **Y.X.:** Writing - review and editing, Writing - original draft, Supervision. **X.M.:** Writing - review and editing, Writing - original draft, Supervision, Resources, Data Curation, Funding acquisition. **L.F.:** Writing - review and editing, Writing - original draft, Validation, Supervision, Resources, Methodology, Investigation, Funding acquisition.

## Ethics declarations

### Competing interests

**Yun Kwok Wing** received a consultation fee from Eisai Co., Ltd., an honorarium from Eisai Hong Kong for a lecture, travel support from Lundbeck HK limited for the overseas conference, and an honorarium from Aculys Pharma, Inc for a lecture.

A pending patent application with invention titled "基于电子病历的模拟临床会诊大语言模型评估方法及系统" / "Method and System for Evaluating Large Language Models in Simulated Clinical Consultations Based on Electronic Health Records" with Xin Ma, Jiahui Niu, and Yibin Li as inventors and Shandong University as the patent owner is received by the China National Intellectual Property Administration under application number 202610852823X. This patent application covers the EHR-grounded AI patient simulation system, simulated clinical consultation workflow, and multidimensional LLM evaluation framework described in this manuscript.

# Supplementary Information

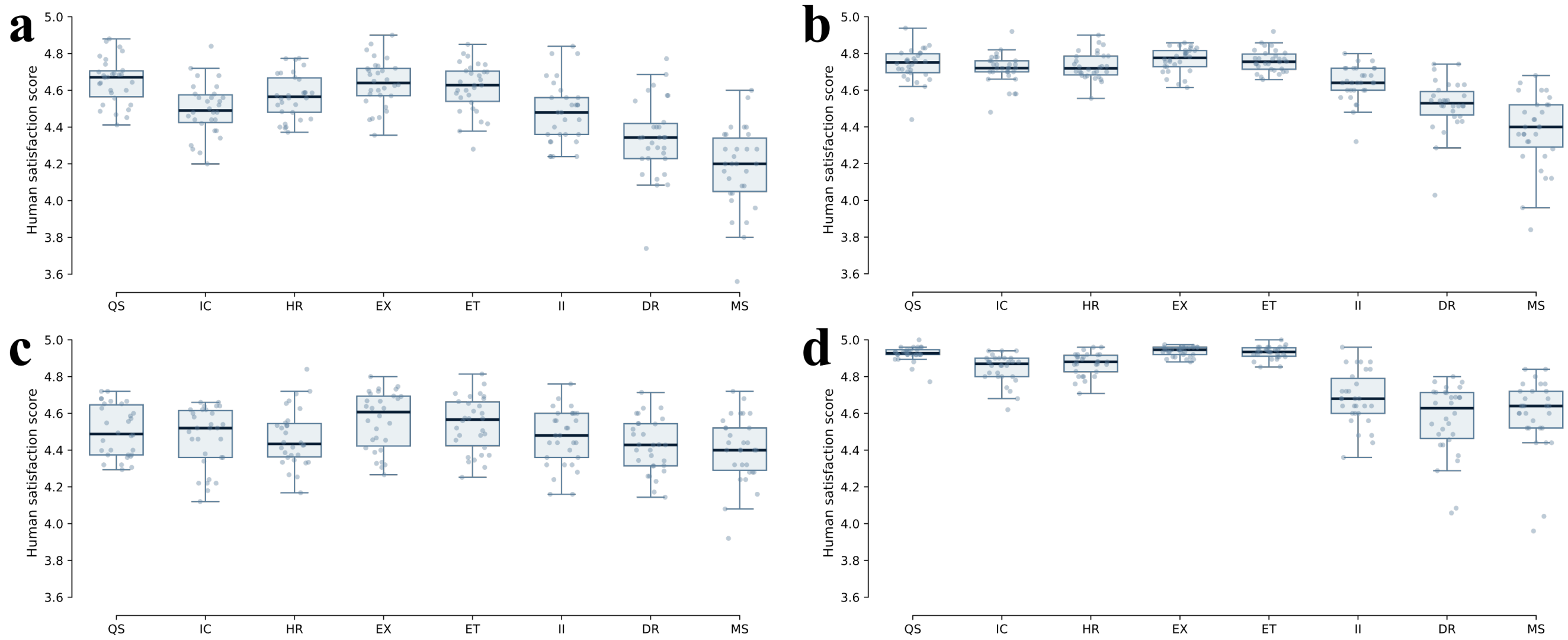

**Supplementary Fig. 1 | Human evaluation of automated clinical assessment across models.** Box plots show case-level mean Likert scores across eight evaluation dimensions for four large language models: **a**, GPT-4o shows consistently high median scores across interaction-related dimensions (e.g., QS = 4.67, EX = 4.64), with greater variability in medication safety. **b**, Claude 4.0 Sonnet exhibits high median scores across all dimensions (e.g., QS = 4.75, EX = 4.78) with tightly concentrated distributions. **c**, Qwen3 maintains generally high median scores (e.g., EX = 4.61, ET = 4.57) but shows broader distributions, particularly in questioning and information collection. **d**, MedGemma achieves near-ceiling median scores across most dimensions (e.g., QS = 4.93, EX = 4.95) with highly concentrated distributions.

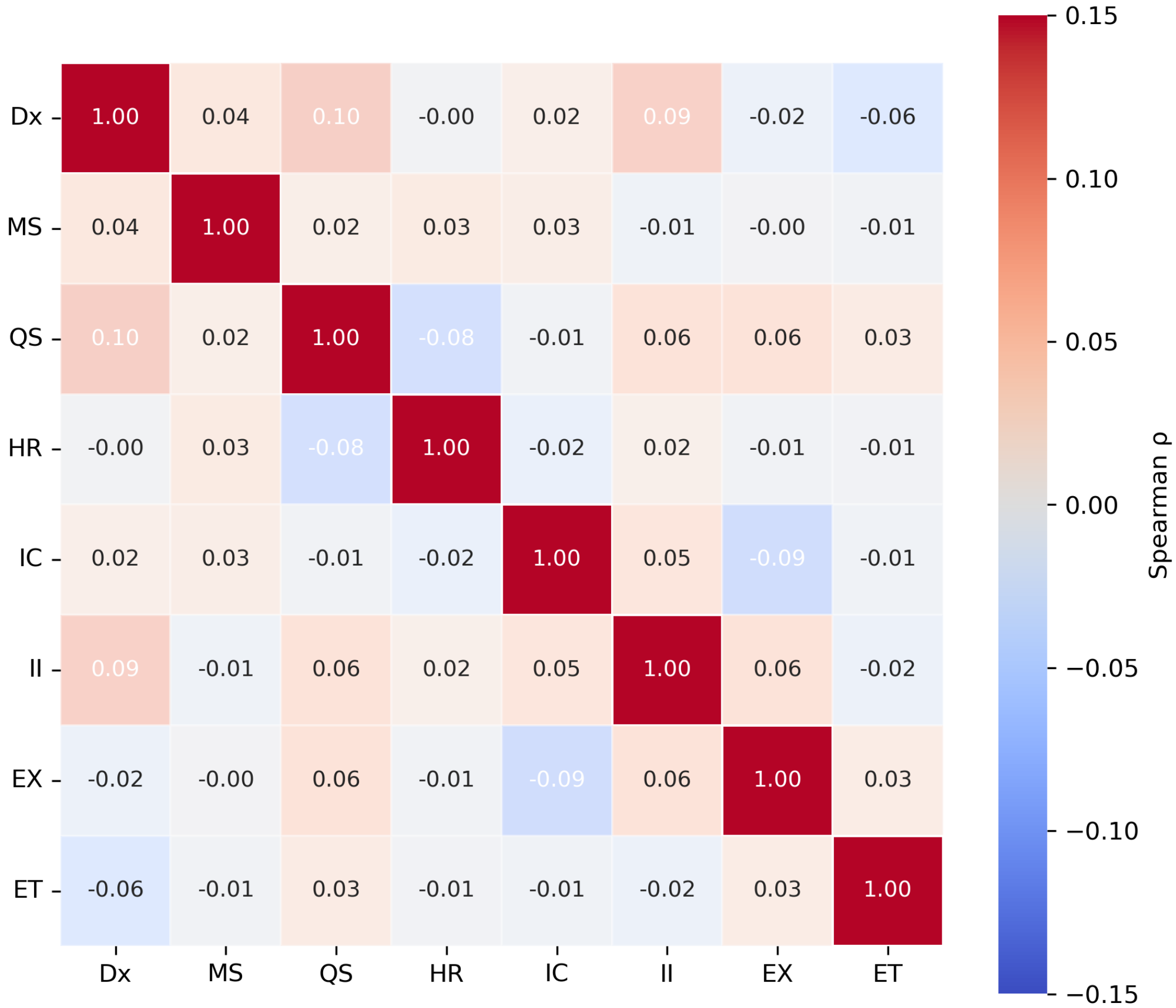

**Supplementary Fig. 2 | Pairwise spearman correlation heatmap among the eight evaluation dimensions**. Dimension names are abbreviated for clarity, and numerical values denote Spearman correlation coefficients. Correlations between dimensions were uniformly weak, with all absolute values below 0.13 and no moderate or strong associations observed, indicating that each dimension captures a distinct aspect of model behavior and supports their use as complementary components of the evaluation framework.

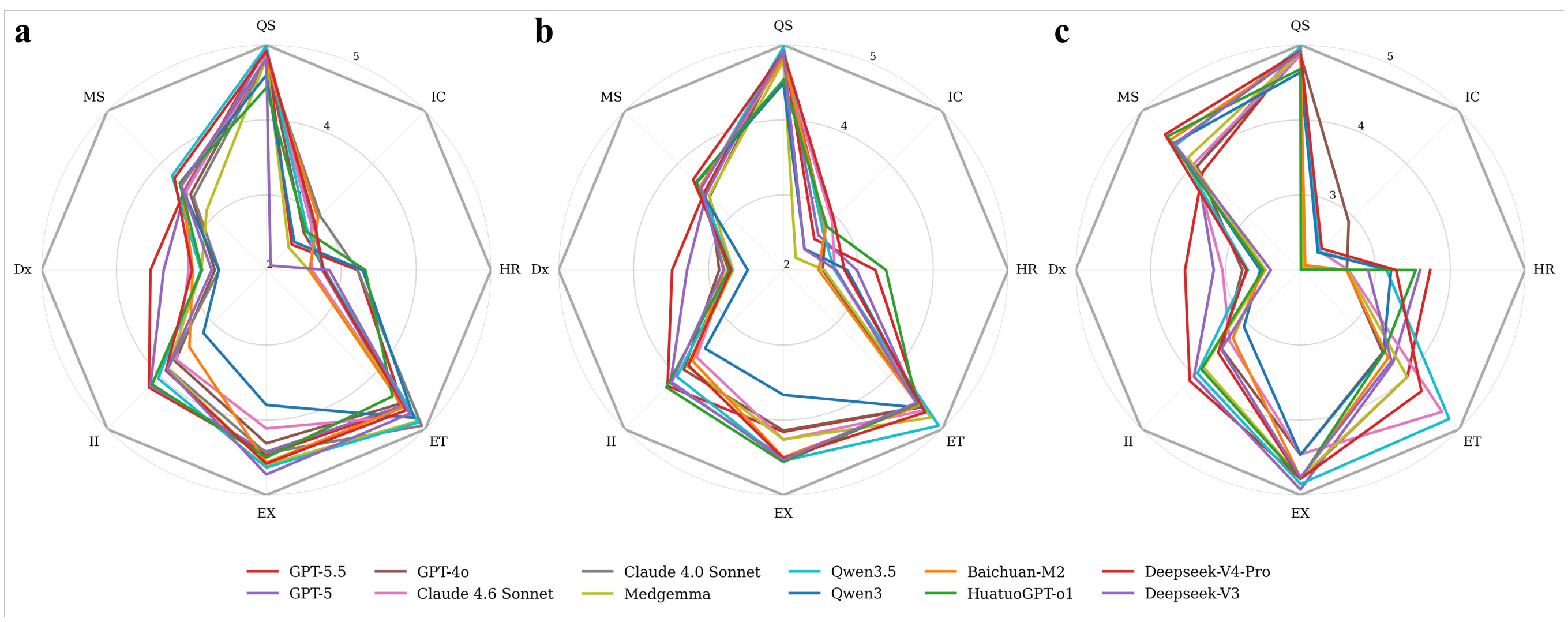

**Supplementary Fig. 3 | Comparative performance of LLMs across clinical evaluation dimensions on the primary CCQA dataset and external validation datasets.** **a**, CCQA dataset. **b**, PMC-Patients dataset. **c**, PCI dataset.

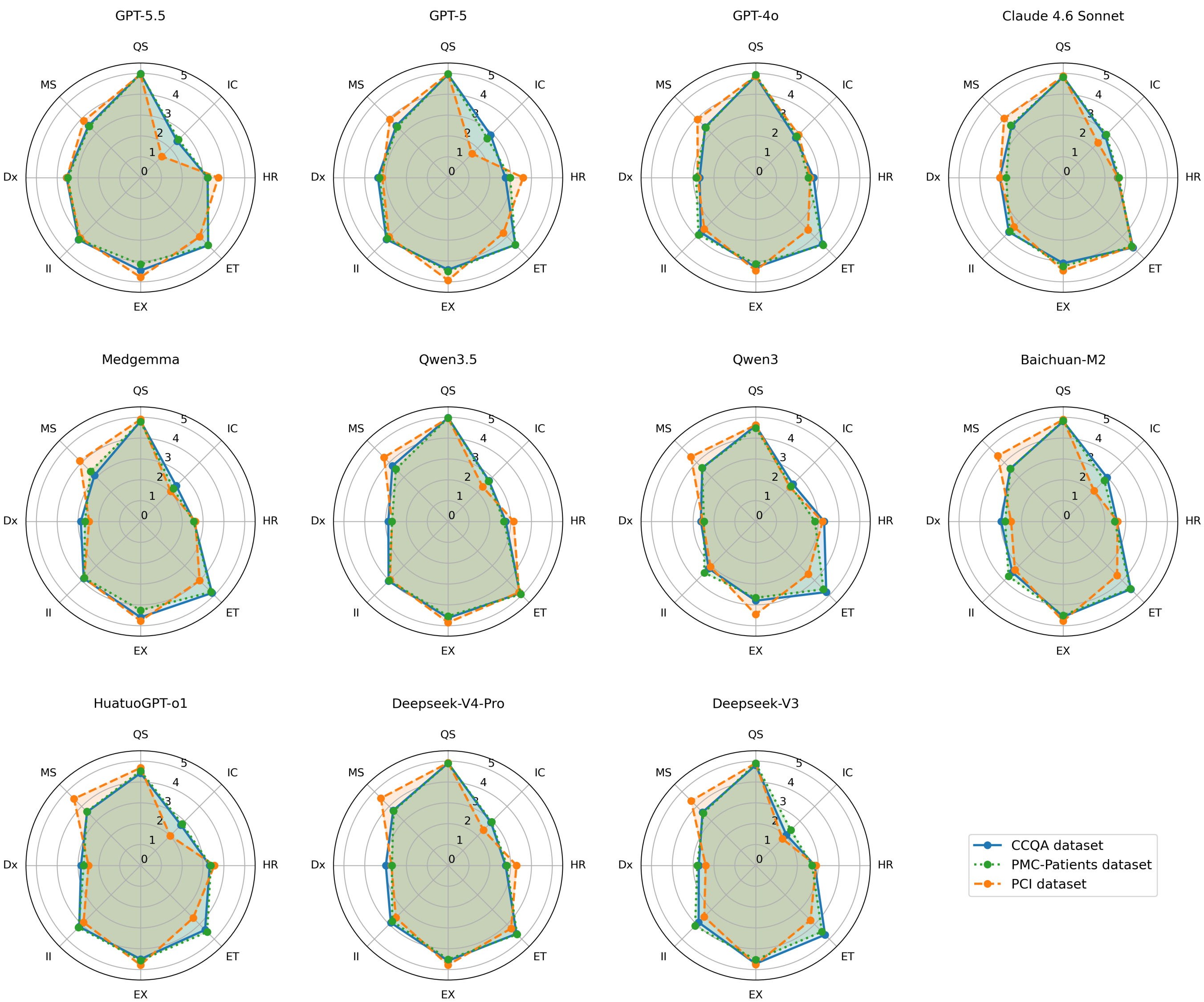

**Supplementary Fig. 4 | Model-specific capability comparison of LLMs across the primary CCQA dataset and external validation datasets**. Model-level capability profiles were largely consistent between CCQA and PMC-Patients but shifted systematically on PCI. Most models showed lower IC scores on PCI, except GPT-4o, while ET scores increased for most models except Claude 4.6 Sonnet and Qwen3.5. MS scores were also elevated across the majority of models.

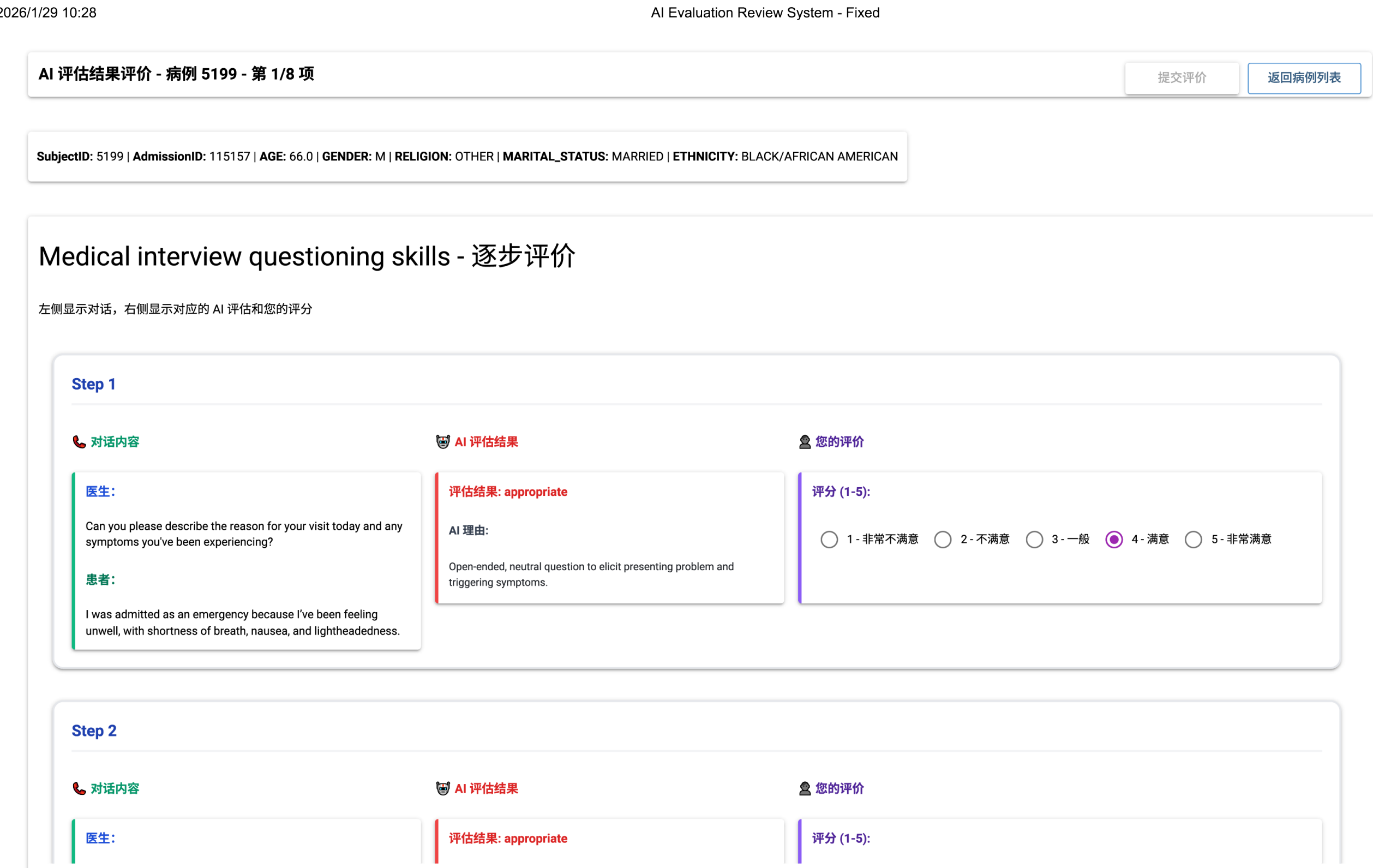

**Supplementary Fig. 5 | Interface for step-wise human evaluation of AI outputs.** Screenshot of the web-based evaluation interface used for human annotation. For each dialogue step, the AI-generated assessment is displayed alongside the original interaction, and human evaluators are asked to rate the appropriateness of the AI's judgment on a Likert scale.

**Supplementary Table 8 | Median (IQR) of human satisfaction scores across evaluation dimensions for different models.** Scores ranged from 4.20 to 4.95.

| **Model** | QS | IC | HR | EX | ET | II | Dx | MS |
|---|---|---|---|---|---|---|---|---|
| GPT-4o | 4.67(0.141) | 4.49(0.150) | 4.57(0.187) | 4.64(0.149) | 4.63(0.164) | 4.48(0.200) | 4.34(0.191) | **4.20(0.290)** |
| Claude 4.0 Sonnet | 4.75(0.103) | 4.72(0.060) | 4.72(0.102) | 4.78(0.087) | 4.75(0.082) | 4.64(0.120) | 4.53(0.127) | 4.40(0.230) |
| Qwen3 | 4.49(0.272) | 4.52(0.255) | 4.43(0.181) | 4.61(0.271) | 4.57(0.238) | 4.48(0.240) | 4.43(0.229) | 4.40(0.230) |
| MedGemma | 4.93(0.026) | 4.87(0.100) | 4.88(0.089) | **4.95(0.040)** | 4.93(0.045) | 4.68(0.190) | 4.63(0.250) | 4.64(0.200) |

**Supplementary Table 2 | Additional demographic characteristics of the CCQA cohort.**

| **Characteristic** | **Count (N) / Value** | **Proportion (%)** |
|---|---|---|
| *Religion* | | |
| Christian | 262 | 59.95 |
| Other | 127 | 29.06 |
| Non-Christian | 48 | 10.98 |
| *Marital Status* | | |
| Married/Partnered | 183 | 41.88 |
| Never Married/Single | 140 | 32.03 |
| Divorced/Separated | 101 | 23.11 |
| Other | 3 | 0.69 |

**Supplementary Table 3 | Clinical consultation quality evaluation metrics.** Multiple failure patterns are defined for each evaluation dimension, with predefined penalty scores assigned to each pattern. Corresponding deductions are applied upon the occurrence of a failure pattern, subject to an upper limit on the total penalty within each dimension.

| **Evaluation dimensions** | **Failure patterns** | **Definition/Rule** | **Per** | **Max** |
|---|---|---|---|---|
| Medical interview questioning skills | Logical gap | Abrupt topic shifts without clear connection to prior content. | 0.5 | 1.0 |
| | Repetitive question | Repeated questions requesting already provided medical information. Clarifying vague answers or exploring new details is not considered repetition. | 0.5 | 1.0 |
| | Unclear question | Question combining two or more unrelated clinical topics in a single unstructured sen-tence without clear boundaries. | 0.5 | 1.5 |
| | Suggestive question | Question that suggest or bias the patient's answer. | 0.25 | 0.5 |
| | Confrontational question | Question that sound aggressive, judgmental, or defensive-provoking. | 0.25 | 1.0 |
| Information coverage | Major omission | A critical domain is entirely missing, undermining clinical understanding. | 1.0 | 3.0 |
| | Minor omission | A contextually relevant domain is not addressed, reducing completeness. | 0.5 | 2.0 |
| Handling of ambiguous patient responses | Failure to clarify unclear expressions | Failure to clarify vague or grouped symptom descriptions within three turns. | 0.5 | 2.5 |
| | Failure to address uncertain answers | Failure to address "don't know / not sure" responses with clarification or follow-up within three turns. | 0.5 | 2.5 |
| Ethical and professional conduct | Disrespectful tone | Rude, dismissive, or condescending communication. | 1.0 | 2.0 |
| | Bias or assumption | Unjustified assumptions about lifestyle, behavior, or illness cause. | 1.0 | 2.0 |
| | Privacy violation | Overly intrusive questions without medical justification. | 1.0 | 2.0 |
| | Lack of empathy | Ignoring or downplaying patient concerns or emotions. | 1.0 | 2.0 |
| Clarity and transparency od clinical explanations | Major clarity failure | Use of unexplained medical terminology that seriously hinders patient understanding. | 1.0 | 2.0 |
| | Minor clarity failure | Use of potentially confusing terms without clarification, causing mild comprehension difficulty. | 0.5 | 1.5 |
| | Lack context for question | Asking potentially alarming or confusing questions without explaining purpose ore relevance. | 0.5 | 1.5 |
| Information intergration | Major omission | Missing critical information affecting diagnosis or treatment. | 0.5 | 1.5 |
| | Minor omission | Missing contextual information not affecting key clinical decisions. | 0.1 | 0.5 |
| | Major fabrication | False or unsupported information that could mislead diagnosis or treatment. | 1.5 | 3.0 |
| | Minor fabrication | Interpretive errors not materially affecting clinical decisions. | 0.5 | 0.5 |
| Diagnostic accuracy and reasoning | Exact match | Final diagnosis exactly matches any reference diagnosis. | 0.0 | 0.0 |
| | Partial match | Diagnosis differs in wording or subtype but represents the same clinical concept. | 0.5 | 0.5 |
| | Minorly incorrect | Diagnosis is plausible but incomplete or misses key aspects. | 1.0 | 1.0 |
| | Majorly incorrect | Diagnosis is unrelated or misleading. | 1.5 | 1.5 |
| | Information omission | Failure to include key patient information in reasoning. | 0.25 | 0.5 |
| | Incorrect assumption | Reasoning introduces unsupported facts treated or confirmed. | 0.5 | 0.5 |
| | Flawed reasoning process | Clinically invalid, disorganized, or step-jumping reasoning. | 0.25 | 0.5 |
| | Diagnosis reasoning mismatch | The reasoning process does not logically lead to the clinician's | 0.5 | 0.5 |
| | | final diagnosis. | | |
| | Diagnostic questiong bias | Assign this when multiple reasonable differential diagnoses exist, but only one line of questioning is pursued without justification. | 0.5 | 0.5 |
| | Correct diagnoses can be made from the conversation | If the dialogue provides sufficient logical evidence to support at least one reference diagnosis, label as true; otherwise, label as false. | - | - |
| Medication safety and justification | Reference drug deviation | Drug is not in reference list and is not a reasonable or safe alternative. | 0.3 | 1.0 |
| | Drug diagonsis mismatch | Drug has no clear therapeutic relevance to the diagnosis or diagnostic reasoning. | 0.5 | 1.0 |
| | Unsafe drug use | Drug is contraindicated or potentially harmful given known patient factors. | 0.5 | 1.0 |
| | Missing important drugs | Key reference medication is entirely omitted. | 0.3 | 1.0 |
| | Incorrect or missing explanation | Drug rationale is unclear, inaccurate, or missing, regardless of drug appropriate-ness. | 0.3 | 1.0 |

**Supplementary Table 4 | Demographic characteristics of the PMC-Patients external validation cohort.**

| Characteristic | Count (N) / Value | Proportion (%) |
| --- | --- | --- |
| **Total Patients** | 119 | - |
| **Total Cases** | 119 | - |
| **Age, years (mean ± SD)** | 48.86 ± 18.28 | - |
| **Female (F)** | 54 | 45.38 |
| **Male (M)** | 65 | 54.62 |

**Supplementary Table 5 | Demographic characteristics of the PCI external validation cohort.**

| Characteristic | Count (N) / Value | Proportion (%) |
|---|---|---|
| **Total Patients** | 67 | - |
| **Total Admissions** | 67 | - |
| **Age, years (mean ± SD)** | 19.30 ± 12.21 | - |
| **Female (F)** | 42 | 62.69 |
| **Male (M)** | 25 | 37.31 |

**Supplementary Note 1 | Outcome-level evaluation.**

To evaluate outcome-level performance, we consider both diagnostic accuracy and medication quality. Let $N$ denote the total number of cases. For each case i, let $y_i$ denote the reference and predicted diagnoses, respectively, and let $R_i$ and $G_i$ denote the sets of recommended and reference medications.

The outcome-level diagnostic accuracy is defined as

$$\text{Accuracy} = \frac{1}{N}\sum_{i=1}^{N} I(\hat{y}_i \in \mathcal{M}(y_i)), \tag{1}$$

where $I(\cdot)$ is the indicator and $\mathcal{M}(y_i)$ denotes the set of acceptable diagnoses, including both exact and clinically valid partial match.

For medication evaluation, let $FP_i$ denote the number of incorrectly recommended medications, $FN_i$ is the number of missing essential medications, $U_i$ is the number of unsafe medication usages, and $D_i$ is the number of drug--diagnosis mismatches. The medication F1 score is computed at the case level as

$$\text{Precision}_i = \frac{TP_i}{TP_i + FP_i}, \quad \text{Recall}_i = \frac{TP_i}{TP_i + FN_i}, \tag{2}$$

where

$$TP_i = \max(|R_i| - FP_i, 0), \tag{3}$$

The case-level F1 score is defined as

$$F1_i = \frac{2 \cdot \text{Precision}_i \cdot \text{Recall}_i}{\text{Precision}_i + \text{Recall}_i}, \tag{4}$$

and the overall score is given by

$$\text{avg_F1} = \frac{1}{N}\sum_{i=1}^{N} F1_i, \tag{5}$$

The unsafe drug use rate is defined as

$$\text{Unsafe Rate} = \frac{1}{N}\sum_{i=1}^{N} I(U_i > 0), \tag{6}$$

and the drug–diagnosis mismatch error rate is computed as

$$E_i = \frac{D_i}{\max(|R_i|, 1)}, \text{Mismatch Rate} = \frac{1}{N}\sum_{i=1}^{N} E_i. \tag{7}$$

**Supplementary Note 2 | Standardized clinical consultation prompt (LLMs).**

You will simulate an experienced professional clinician and conduct a structured, human-like medical consultation. The consultation is taking place in the outpatient clinic. You and the patient are sitting face to face in the consultation room, engaging in a natural doctor-patient conversation.

Your task is to conduct the consultation in a structured, human-like way, as if you were the treating physician. Your behavior must reflect the style of a real doctor, avoid letting patients know that you are AI, and communicate like a doctor in a real clinic. Ask only one question at a time, do not ask multiple questions at a time or break down the questions.

The consultation must strictly end within 15 turns. If the 15th turn is reached, you must stop and give the final diagnosis and treatment plan, even if information is incomplete. The question should be concise, logical, and contain moderate information.

The current situation is as follows: Patient Information: {initial_info}, Conversation History: {conversation_history} You are currently at turn {turn_number} of a maximum of 15 turns.

Your tasks: 1. Read the conversation history carefully to determine whether more information is needed. 2. If necessary, ask the next natural question to move the consultation forward. 3. You must strictly end the consultation no later than the 15th turn. - If sufficient information is available earlier, end before the 15th turn. - If the 15th turn is reached, you must end immediately, even if information is incomplete. Provide a clear and precise diagnosis, and issue a complete, justified prescription list. The prescription must include specific medication names. Each drug must be clearly linked to the patient's specific clinical issues—including the primary disease, complications, symptoms, and risk prevention. Do not omit treatment for any relevant problem. Ensure the medication plan is comprehensive, safe, and clinically appropriate. 4. Record all clinical information about the patient mentioned in the conversation. For any missing information, explicitly state "Not available".

Please respond using only one of the two JSON formats below. Do not include any additional text or markdown formatting.

Next question format:

{ type: question, content: <Your next natural continuation or prompt in the interview>}Final diagnosis and treatment plan answer format: { type: answer, content: { diagnosis: <Final diagnosis>, diagnosis_reasoning: <Detailed step-by-step diagnostic reasoning, explaining how this diagnosis was reached>, recommended_drugs: [DRUG1, DRUG2, ...], drug_reasoning: { DRUG1: <Why this drug is appropriate for the patient>, DRUG2: <...> }, category_summaries: { symptoms: <...>, medical_history: <...>, family_history: <...>, vital_signs: <...>, allergies: <...> } }}